\documentclass[11pt]{article}

\usepackage{color}
\usepackage{amsmath}
\usepackage{amssymb}
\usepackage{fullpage}
\usepackage{graphicx}
\usepackage{dsfont}
\usepackage{algorithm}
\usepackage{setspace}
\usepackage{appendix}
\usepackage{comment}
\usepackage{multirow}
\usepackage{hyperref}
\usepackage{authblk}
\usepackage{xr}
\newcommand{\bs}{\boldsymbol}
\makeatletter
\newcommand{\vast}{\bBigg@{4}}
\newcommand{\Vast}{\bBigg@{5}}
\makeatother

\allowdisplaybreaks

\usepackage{natbib}
 \bibpunct[, ]{(}{)}{,}{a}{}{,}%

\newcommand{\CH}[1]{{\color{black} #1}}   

\newtheorem{theorem}{Theorem}
\newtheorem{lemma}{Lemma}
\newtheorem{corollary}{Corollary}
\newtheorem{remark}{Remark}
\newtheorem{proposition}{Proposition}

\begin{document}
\title{Adaptive Policies for Perimeter Surveillance Problems}
\author[1]{James A. Grant\thanks{j.grant@lancaster.ac.uk; corresponding author}}
\author[2]{David S. Leslie\thanks{d.leslie@lancaster.ac.uk}}
\author[3]{Kevin Glazebrook\thanks{k.glazebrook@lancaster.ac.uk}}
\author[4]{Roberto Szechtman\thanks{rszechtm@nps.edu}}
\author[3]{Adam N. Letchford\thanks{a.n.letchford@lancaster.ac.uk}}
\affil[1]{STOR-i Centre for Doctoral Training, Lancaster University, UK}
\affil[2]{Department of Mathematics and Statistics, Lancaster University, UK}
\affil[3]{Department of Management Science, Lancaster University, UK}
\affil[4]{Department of Operations Research, Naval Postgraduate School, CA, USA}
\maketitle

\onehalfspacing

\abstract{We consider the problem of sequentially choosing observation regions along a line, with an aim of maximising the detection of events of interest. Such a problem may arise when monitoring the movements of endangered or migratory species, detecting crossings of a border, policing activities at sea, and in many other settings. In each case, the key operational challenge is to learn an allocation of surveillance resources which maximises successful detection of events of interest.  We present a combinatorial multi-armed bandit model with Poisson rewards and a novel filtered feedback mechanism - arising from the failure to detect certain intrusions - where reward distributions are dependent on the actions selected. Our solution method is an upper confidence bound approach and we derive upper and lower bounds on its expected performance. We prove that the gap between these bounds is of constant order, and demonstrate empirically that our approach is more reliable in simulated problems than competing algorithms.}%


\section{Introduction} \label{sec::intro}
Many common surveillance tasks concern the detection of activity along a border or perimeter. Monitoring the movements of endangered or migratory species through crossings using camera traps, covertly tracking illegal fishing in territorial waters via adaptive satellite technology, and quantifying traffic across a border using drone technology are a few among many examples of important potential aims in this domain. Equally, a number of common scheduling challenges involve events arising through time. For instance, scheduling call center staff to meet random arrivals, or deciding what times traffic cameras should be in operation to catch speeding drivers. 

Approaches to the optimal design of observation strategies are invaluable not only at the operational level, but also at the strategic level because they can inform decision makers about expected outcomes for different budget scenarios and policies.  In each of these tasks the notion of optimality can be equated to maximising the rate of detection of events, or equivalently, detecting as many events as possible over some fixed time horizon. 

We consider a scenario where observations are made by a team of \emph{searchers} (representing cameras, sensors, human searchers, etc.), coordinated by a central agent referred to as the \emph{controller} who chooses which segments of a line segment each searcher will observe. As the line segment may be thought of as indexing space or time, the formulation captures a wide range of examples (we will discuss the spatial problem in what follows for ease of exposition). We will assume that events arise according to a Poisson process and the likelihood of an event being detected depends on the allocation of resource chosen by the controller. 

The problem of designing an optimal deployment of searchers becomes truly challenging when the number of available searchers is insufficient to guarantee perfect detection of all events, which is often the case in tight fiscal environments. In such a setting the controller faces a classic \emph{resource allocation problem}, where the action set is the set of possible allocations of searchers to segments of the line and the controller aims to find an action which maximises the rate of detection. To compute this rate of detection the controller must know the rate at which events occur along the length of the line and the probabilities with which searchers detect events that have appeared at particular points (under a particular allocation of searchers to parts of the line). It is, of course, a strong assumption that such information is available, particularly at the beginning of a new project. 

In this work we consider the more realistic setting where the rate at which events occur is unknown. When the rate at which events occur is unknown the controller has two broad options: 
\begin{enumerate}
\item[(a)] to select an allocation which performs best in expectation according to some prior information (if it exists) and stick to that,
\item[(b)] (if possible) to take an adaptive strategy, which alters the allocation of searchers as observations are collected.
\end{enumerate}
In this second scenario a \emph{sequential resource allocation problem} is faced - where the controller wishes to quickly and confidently converge on an optimal allocation while also ensuring appropriate experimentation.  This sequential problem is our principal concern in this paper. 

To permit analysis of this problem we shall assume two discretisations to simplify the controller's action set. We will consider that opportunities to update the allocation of searchers occur only at particular time points $t \in \mathbb{N}$. Thus, the problem can be thought of as taking place over a series of rounds. We will also suppose that the search space has been divided into a number of cells such that each searcher is allocated a connected set of cells in which to patrol, disjoint from those allocated to other searchers. Imposing this discrete structure on the problem is useful as it allows us to draw on a large literature concerning \emph{multi-armed bandit problems} when designing and analysing solutions to the problem.

Multi-armed bandit problems are relevant to this sequential resource allocation problem because they provide a framework for studying \emph{exploration-exploitation dilemmas}, which is the principal challenge faced by the controller here. In order to reliably make optimal actions, data must be collected from all cells to accurately estimate the expected number of detections associated with an action - i.e. the action space should be \emph{explored}. However, data is being collected on a \emph{live} problem - real events are passing undetected when sub-optimal actions are played. As such there is a pressure to \emph{exploit} information that has been collected and select actions which are believed to yield high detection rates over those with more exploratory value. A balance must be struck. One may suppose that this is a trivial issue which can be resolved by simply searching in all cells in all rounds. However, searching more cells will not necessarily lead to more accurate information or a higher detection rate. Searchers become less effective at detecting events the more cells they are allocated, because events may be undetected if a searcher is aiming to detect over too large a region. Indeed, an optimal action may well be to assign each searcher to just a single cell.

\subsection{Related Literature} \label{subsec::related_lit}
We select a Poisson process as the data generating model for our problem. The Poisson process is famously widely used as a model for spatial and spatiotemporal event data in many settings, such as ecology \citep{Heikkinen1999,Serra2014}, and arrival process modelling \citep{Benes1957,Weinberg2007}. There is a large literature on inference for Poisson processes, which has lead to a variety of sophisticated techniques, such as those involving Gaussian processes \citep{Adams2009,John2018} or kernel-based smoothing \citep{Diggle1985}. However the theoretical properties of the more complex methods are typically only understood asymptotically \citep{Helmers2005,Kirichenko2015, Gugushvili2018} and therefore in the interest of developing tight guarantees on the performance of sequential decision making algorithms, we favour a simple piecewise-constant model for the Poisson process rate in this paper.

Search theory has its origins in WWII with the study of barrier patrols during the Battle of the Atlantic (\cite{koopman1946search}). The works  of \cite{Stone1976} and \cite{Washburn2002} present a much broader and more contemporary range of applications in search theory and detection, and are by now the classic references on the subject. More closely related to our work is \cite{Szechtman2008}, who study the perimeter protection problem when the parameters of the arrival process are fully known, for mobile and fixed searchers. \cite{Carlsson2013} study the problem of optimally partitioning a space in $\mathbb{R}^2$ to maximise a function of an intensity of events over the space. Their problem bears resemblance to the full information version of our problem though our solution method is quite different due to our discretisation of the problem. Our work is, to the best of our knowledge, the first to tackle the learning aspect of such a problem. 

The sequential problem we consider is structurally similar to a combinatorial multi-armed bandit (CMAB) problem \citep{Chen2013}. To permit discussion of a CMAB we first describe the simpler multi-armed bandit (MAB) problem (first attributed to \cite{Thompson1933}), which is a special case. The (stochastic) MAB problem models a scenario where an agent is faced with a series of potential actions (or arms), each associated with some underlying probability distribution. In each of a series of rounds, the agent selects a single action and receives a \emph{reward} drawn from the underlying distribution associated with the selected action. The agent's aim is to maximise her cumulative expected reward over some number of rounds, or equivalently minimise her cumulative \emph{regret} - defined as the difference in expected reward between optimal actions and actions actually selected. To succeed in this the agent must manage an exploration-exploitation trade-off as she learns which actions have high expected reward. 

The CMAB problem models a richer framework where the agent may select multiple actions in each round and her reward is a function of the observations from the underlying distributions associated with the selected actions. \cite{Chen2013} consider a setting where this function may be non-linear. Numerous authors (\cite{Anantharam1987i}, \cite{Gai2012}, \cite{Kveton2015tight}, \cite{Combes2015}, and \cite{Luedtke2016}) consider a special case (known as a \emph{multiple play bandit}) where the reward is simply a sum of the random observations and the number of actions which may be selected in one round is limited. A number of other works have since extended the framework of \cite{Chen2013} to model other novel features.
\cite{Chen2016a} and \cite{Kveton2015cascading} consider a setting where playing a subset of arms may randomly trigger additional rewards from other arms, and \cite{Chen2016b} considers a broader set of non-linear reward functions. However the CMAB model and UCB approach of \cite{Chen2013} is the work closest to ours as the later developments model features that are not present in our setting. The fundamental differences between our model and theirs are that we consider heavy tailed rewards and a setting where reward distributions depend on the selected action.

Reward maximisation in a CMAB problem requires addressing a similar trade-off between exploration and exploitation to that faced in the MAB problem. For MAB-type problems, it has famously been shown that under certain assumptions optimal policies can be derived by formulating the problem as a Markov Decision Process and using an index approach \citep{Gittins2011}. In CMAB problems however, these approaches are inappropriate, not least, since the combinatorial action sets induce dependencies between rewards generated by distinct actions which invalidates Gittins' theory. See also Remark 1 in Section 2. More recently, so called upper confidence bound (UCB) algorithms, first proposed by \cite{Lai1985asymptotically} and \cite{Burnetas1996optimal}, and popularised by \cite{Auer2002}, have attracted much attention as approaches that enjoy efficient implementation and strong theoretical guarantees. These heuristic methods balance exploration and exploitation by selecting actions based on optimistic estimates of the associated expected rewards and can be applied to both MAB and CMAB problems. 

\cite{Auer2002} originally proposed a UCB approach for MAB problems with underlying distributions whose support lies entirely within $[0,1]$. \cite{Chen2013} extended the principles of this algorithm to a version suitable for CMAB problems with nonlinear rewards. Broader classes of unbounded distributions have been considered by other authors. \cite{Cowan2015Normal}, \cite{Bubeck2012}, \cite{Bubeck2013}, and \cite{Lattimore2017} give UCB algorithms suitable for use with unbounded distributions, studying distributions that are Gaussian, have sub-Gaussian tails, known variance and known kurtosis respectively. \cite{Luedtke2016} have studied multiple-play bandits with exponential family distributions. However for CMAB problems with non-linear reward functions attention has focussed on the $[0,1]$ case. Accompanying each of these proposals of UCB algorithms is a corresponding proof which demonstrates the performance of that algorithm achieves the optimal order, albeit with a sub-optimal coefficient.

Stronger performance guarantees (i.e. those with improved leading-order coefficients) have been obtained in MAB problems using Thompson Sampling (TS) type approaches \citep{Kaufmann2012thompson,Agrawal2012,Russo2016,Wang2018} and approaches which utilise the KL divergence of the reward distributions  \citep{Cappe2013,Kaufmann2016}. \cite{Combes2015} have successfully extended the KL divergence based results to multiple play bandits with bounded rewards. However extending these results to the framework of our problem presents a significant analytical challenge, \CH{as the existing theory around KL-based UCB indices relies on independence of the reward generation and action selection mechanisms. Therefore in this work we focus on the theoretical analysis of a more traditional} UCB type approach. A TS alternative is presented and evaluated numerically in Section \ref{sec::numerics}.

\subsection{Key Contributions}
This work makes a number of contributions to the theory of multi-armed bandits and broader online optimisation. Simultaneously, we give a practically useful solution to a real problem encountered in many applications. We summarise the headline contributions below:
\begin{itemize}
\item Introduction of a formal model for sequential event detection problems and an efficient integer programming solution to the full-information version of the problem;
\item Introduction of the \emph{filtered feedback} model for combinatorial multi-armed bandits;
\item Development of a bespoke treatment of combinatorial bandits with \emph{Poisson} rewards, leading to a new martingale inequality for filtered Poisson data and an accompanying UCB approach;
\item Regret analysis yielding an optimal order upper bound on finite time regret of the UCB algorithm and a problem-specific lower bound on asymptotic regret for any uniformly good algorithm.
\end{itemize}
We also present extensive numerical work which displays the robustness of the UCB approach in contrast to its competitors.

\subsection{Paper Outline}
The remainder of the paper is structured as follows. Section \ref{subsec::model} introduces a model of the sequential problem. In Section \ref{sec::full_info} we solve the full information problem (the non-sequential resource allocation problem where the rate function of the arrival process is known). The proposed integer programming solution forms the backbone of the proposed solution methods for the sequential problem. In Section \ref{sec::seq_learn} we introduce a solution method, the \emph{Filtered Poisson Combinatorial Upper Confidence Bound} algorithm, for the sequential resource allocation problem,  and derive a performance guarantee in the form of an upper bound on expected regret of the policy. Here, we also derive a lower bound on the expected regret possible for any policy and thus show that our algorithm has a bound of the correct order. We conclude in Sections \ref{sec::numerics} and \ref{sec::discussion} with numerical experiments and a discussion respectively. 

\section{The Model} \label{subsec::model}

Before introducing solution methods we give a mathematical model of the problem. Throughout the paper, for a positive integer $W$ let the notation $[W]$ represent the set $\{1,2,...,W\}$.

The observation domain (line) comprises $K$ cells which can be searched by $U$ searchers. We write \begin{displaymath}
a_k = u, \quad k \in [K], \enspace u\in [U]
\end{displaymath}
to denote the deployment of searcher $u$ to cell $k$, while \begin{displaymath}
a_k =0, \quad k \in [K]
\end{displaymath}
is used when cell $k$ goes unsearched. An \emph{action} \CH{$\mathbf{a} := (a_1,a_2,...,a_K) \in \{0,1,...,U\}^K$} describes a deployment of the searchers across the line. We impose the requirement that $\mathbf{a} \in \mathcal{A}$, the \emph{action set}, where \begin{displaymath}
\mathcal{A}=\{\mathbf{a}  : \enspace a_i=a_j=u  \Rightarrow a_k=u, \enspace \forall i,j,k \in [K] :  i \leq k \leq j, \enspace i <j, \text{ and } \forall u \in [U]\}.
\end{displaymath} These conditions on $\mathcal{A}$ ensure that searchers are assigned to disjoint connected sub-regions of the perimeter. The actions are uniquely defined by indicator variables $a_{iju} \in \{0,1\}$ for $i,j \in [K]$, $i<j$ and $u \in [U]$ such that \begin{displaymath}
a_{iju}=1 \Leftrightarrow \text{agent } u \text{ is assigned to the cells } \{i,i+1,...,j\} \text{ only.}
\end{displaymath}

Each action $\mathbf{a} \in \mathcal{A}$ gives rise to a certain detection probability $\gamma_k(\mathbf{a}) \in [0,1]$ in all cells $k \in [K]$. The detection probabilities capture the effectiveness of each searcher in observing an event in a specific cell. We write $\bs\gamma(\mathbf{a})$ for the $K$-vector whose $k^{th}$ component is $\gamma_k(\mathbf{a})$.
The detection probabilities are structured such that for any $\mathbf{a}, \mathbf{b} \in \mathcal{A}$ and $i \leq j$, \begin{displaymath}
a_{iju}=b_{iju}=1 \Rightarrow \gamma_k(\mathbf{a})=\gamma_k(\mathbf{b}), \enspace \forall k \text{ such that } i \leq k \leq j.
\end{displaymath}
Hence, the detection probability in a cell depends only on the sub-region assigned to the single agent searching that cell and is unaffected by the sub-regions assigned to other searchers. We assume that if a cell is searched there will be some non-zero probability of detecting events that occur. That is to say for any $k \in [K]$, $\gamma_k(\mathbf{a})>0$ for any $\mathbf{a} \in \mathcal{A}$ such that $a_k \neq 0$.

We consider two cases with respect to knowledge of the detection probabilities: \begin{enumerate}
\item[(I)] The detection probabilities $\bs\gamma(\mathbf{a})$ are known for all $\mathbf{a} \in \mathcal{A}$. This scenario occurs when the controller knows $\bs\gamma(\mathbf{a})$ from the past.
\item[(II)] The functions $\bs\gamma$ have a particular known parametric form but unknown parameter values. This case is realistic when properties of the detection probabilities are dictated by physical considerations, such as the searchers' speed, the visibility in particular locations or the time for which an event is observable.
\end{enumerate}

Our sequential decision problem may now be described as follows: \begin{enumerate}
\item At each time $t \in \mathbb{N}$ an action $\mathbf{a}_t \in \mathcal{A}$ is taken, inducing a detection probability $\gamma_k(\mathbf{a}_t)$ in each cell $k \in [K]$;
\item Events are generated by $K$ independent Poisson processes, one for each cell. We use $X_k$ to denote the number of events in cell $k$ (whether observed or not) occurring during the period of a single search. We have \begin{displaymath}
X_k \sim Pois(\lambda_k), \enspace k \in [K]
\end{displaymath}
where the rates $\lambda_k \in \mathbb{R}_{+}$ are unknown, and write $\lambda_{max} \geq \max_{k \in [K]} \lambda_k$ for a known upper bound on the arrival rates. We use $X_{kt}$ for the number of events generated in cell $k$ during search $t$.
\item Should action $\mathbf{a}_t$ be taken at time $t$, a random vector of events $\mathbf{Y}_t = \{Y_{1t},Y_{2t},...,Y_{Kt}\} \in \mathbb{N}^K$ is observed. Events in the underlying $X$-process are observed or not independently of each other. We write \begin{displaymath}
Y_{kt} | X_{kt}, \mathbf{a}_t \sim Bin(X_{kt},\gamma_k(\mathbf{a}_t)), \enspace k \in [K].
\end{displaymath}
It follows from standard theory that \begin{displaymath}
Y_{kt}|\mathbf{a}_t \sim Pois(\lambda_k\gamma_k(\mathbf{a}_t)), \enspace k\in [K],
\end{displaymath}
and are independent random variables. It follows that the mean number of events observed under action $\mathbf{a}$ is given by \begin{displaymath}
r_{\bs\lambda, \bs\gamma}(\mathbf{a}) := \bs\gamma(\mathbf{a})^T\bs\lambda,
\end{displaymath}
where $T$ denotes vector transposition and $\bs\lambda$ is the $K$-vector whose $k^{th}$ component is $\lambda_k$.

\item We write \begin{displaymath}
\mathbf{H}_t=\{\mathbf{a}_1,\mathbf{Y}_1,...,\mathbf{a}_{t-1},\mathbf{Y}_{t-1}\}
\end{displaymath}
for the \emph{history} (of actions taken and events observed) available to the decision-maker at time $t \in \mathbb{N}$. A \emph{policy} is a rule for decision-making and is determined by some collection of functions $\big\{\pi_t: \mathbf{H}_t \rightarrow \mathcal{A}, t \in \mathbb{N} \big\}$ adapted to the filtration induced by $\mathbf{H}_t$. In practice a policy will be determined by some algorithm $A$. We will use the terms policy and algorithm interchangeably in what follows.
\end{enumerate}

The goal of analysis is the elucidation of policies whose performance (as measured by the mean number of events observed) is strong uniformly over $\bs\lambda, \bs\gamma$ and over partial horizons $\{1,2,...,n\} \subseteq \mathbb{N}$. We write \begin{displaymath}
\mathds{E}_A\bigg(\sum_{t=1}^n r_{\bs\lambda,\bs\gamma}(\mathbf{a}_t)\bigg)
\end{displaymath}
for the mean number of events observed up to time $n \in \mathbb{N}$ under algorithm $A$. If we write \begin{displaymath}
\text{opt}_{\bs\lambda,\bs\gamma} := \max_{\mathbf{a} \in \mathcal{A}} r_{\bs\lambda,\bs\gamma}(\mathbf{a}),
\end{displaymath}
then it is plain that, for any choice of $A$ \begin{displaymath}
n \cdot \text{opt}_{\bs\lambda,\bs\gamma} \geq \mathds{E}_A\bigg(\sum_{t=1}^n r_{\bs\lambda,\bs\gamma}(\mathbf{a}_t)\bigg),
\end{displaymath} with achievement of the left hand side dependent on knowledge of $\bs\lambda$. Assessment of algorithms will be based on the associated \emph{regret function}, the expected reward lost through ignorance of $\bs\lambda$, given for algorithm $A$ and horizon $n$ by \begin{equation}
Reg_{\bs\lambda,\bs\gamma}^A(n) := n \cdot \text{opt}_{\bs\lambda,\bs\gamma} - \mathds{E}_A\bigg(\sum_{t=1}^n r_{\bs\lambda,\bs\gamma}(\mathbf{a}_t)\bigg), \label{eq::regret}
\end{equation}
which is necessarily positive and nondecreasing in $n$, for any fixed $A$. In related bandit-type problems the regret of the best algorithms typically grows at $O(\log(n))$ uniformly across all $\bs\lambda$. We will demonstrate both that this is also the case for the algorithms we propose and that the best achievable growth for this problem is also $O(\log(n))$.

\begin{remark}
An alternative, indeed classical, formulation uses Bayes sequential decision
theory. Here the goal of analysis is the determination of an algorithm $A$ to maximise \begin{displaymath}
\mathds{E}_\rho \Bigg[\mathds{E}_A\bigg(\sum_{t=1}^n r_{\bs\lambda,\bs\gamma}(\mathbf{a}_t)\bigg) \Bigg]
\end{displaymath}
where the outer expectation is taken over some prior distribution $\rho $ for the unknown $\bs\lambda$. A standard approach would formulate this as a Markov Decision Process (MDP) with an informational state at time $t$ taken to be some sufficient statistic for $\bs\lambda$. The objections to this approach in this context are many. First, any serious attempt to derive such a formulation which is likely tractable will require strong assumptions on the prior $\rho $ including, for example, independence of the components of $\bs\lambda$. These would each typically have a conjugate gamma prior. Even then the resulting dynamic program would be computationally intractable for any reasonable choices of $K$ and $n.$ Second, the realities of our problem (and, indeed, many others) are such that specification of any reasonably informed prior is impractical. Confidence in the analysis would inevitably require robustness of the performance of any proposed algorithm to specification of the prior. Indeed,
our formulation centred on regret simply seeks robustness of
performance with respect to values of the unknown $\bs\lambda$. Third, the MDP approach would require up front specification of the decision horizon $n.$ This is practically undesirable for our problem. Moreover, the value of $n$ is not unimportant. It will determine the nature of good policies in important ways. For example, the ``last'' decision at time $n$ is guaranteed to be optimally ``greedy'' since there is no further need to learn about $\bs\lambda$ at that point.
\end{remark}

\section{The Full Information Problem} \label{sec::full_info}
In order to develop strongly performing policies, it is critical that we are able to solve the \emph{full information} optimisation problem \begin{displaymath}
\text{opt}_{\bs\lambda,\bs\gamma} := \max_{\mathbf{a} \in \mathcal{A}} r_{\bs\lambda,\bs\gamma}(\mathbf{a})
\end{displaymath}
for any pre-specified $\bs\lambda \in (\mathds{R}_+)^K$. A naive proposal for a policy addressing the problem outlined in the previous section would choose an action $\mathbf{a}_t$ at time $t$ to solve the full information problem for some estimate $\bs\lambda_t$ of the unknown $\bs\lambda$ available at time $t$. While such a proposal would fail to adequately address the challenge of learning about $\bs\lambda$, we will in the succeeding sections develop effective algorithms which choose allocations determined by solutions of full information problems for carefully chosen $\bs\lambda$-values.

A challenge to the solution of the full information problem is the non-linearity in $\mathbf{a}$ of the objective $r_{\bs\lambda,\bs\gamma}(\mathbf{a})$ inherited from the non-linearity of the detection mechanism $\bs\gamma(\mathbf{a})$. To develop efficient solution approaches we produce a formulation as a linear integer program (IP) in which this non-linearity is removed by precomputing key quantities. In particular we write \begin{displaymath}
q_{\bs\lambda,\bs\gamma,iju} = \sum_{k=i}^j \gamma_k(\mathbf{a}_{iju})\lambda_k
\end{displaymath}
for the mean number of events detected when agent $u$ is allocated to the subregion $\{i,i+1,...,j\}$ where $\mathbf{a}_{iju}$ is any $\mathbf{a} \in \mathcal{A}$ such that $a_{iju}=1$. Efficient solution of the full information problem relies on precomputing these $q_{\bs\lambda,\bs\gamma,iju}$ for all $1 \leq i \leq j \leq K$, and $u \in [U]$. We now have that \begin{align}
\text{opt}_{\bs\lambda,\bs\gamma}=\max_{\{a_{iju}, 1\leq i \leq j \leq K, u \in [U]\}}\sum_{i=1}^K \sum_{j=i}^K \sum_{u=1}^U q_{\bs\lambda,\bs\gamma,iju}a_{iju}& \label{eq::IP}\\
\text{such that } \quad \sum_{i=1}^K \sum_{j=i}^K a_{iju} \leq &1, \enspace u \in [U] \nonumber \\
\sum_{i=1}^k \sum_{j=k}^K \sum_{u=1}^U a_{iju} \leq &1, \enspace k \in [K] \nonumber \\
a_{iju} \in &\{0,1\}, \enspace 1 \leq i \leq j \leq K, \enspace u \in [U]. \nonumber
\end{align}
The first constraint above guarantees that each searcher $u$ is assigned to at most one sub-region while the second constraint guarantees that each cell $k$ is searched by at most one searcher. We view the solution of (\ref{eq::IP}) as the optimal allocation strategy and the optimal value function as the best achievable performance for an agent with perfect knowledge of $\bs\gamma$ and $\bs\lambda$.

When we require solutions to the full information problem for the implementation of algorithms for the problem described in the preceding section, we solve an appropriate version of the above IP (ie, for suitably chosen $\bs\lambda$) by means of branch and bound. While it can be shown that the IP (\ref{eq::IP}) belongs to a class of problems which is NP-hard (see Appendix A) we find that the solution of this IP is very efficient in practice. We believe that this is because the solution of the LP-relaxation of (\ref{eq::IP}) often coincides with the exact solution of the IP. Indeed, in empirical tests this occurred more than 90\% of the time and in the remaining instances the gap between the two solutions was always less that 1\%. For all problem sizes considered in this paper the pre-processing and solution steps can be completed in less than a second using basic linear program solvers in the statistical programming language \texttt{R} on a single laptop.

\section{Sequential Problem} \label{sec::seq_learn}
In the sequential problem, the controller's objective is to minimise regret (\ref{eq::regret}) over a sequence of rounds. To do so the controller must construct a strategy which balances exploring all cells to accurately estimate the underlying rate parameters $\bs\lambda$, while also exploiting the information gained to detect as many events as possible. In this section we introduce and analyse two upper confidence bound (UCB) algorithms as policies for the case of fully known detection probabilities (case (I)) and the case where only the nature of the scaling of detection probabilities is known (case (II)).

The model we introduced in Section \ref{subsec::model} is closely related to the \emph{Combinatorial Multi Armed Bandit problem} (CMAB) model of \cite{Chen2013}. The CMAB problem models a scenario where a decision-maker is faced with a set of $K$ basic actions (or \emph{arms}) each associated with a random variable of unknown probability distribution. In each round $t \in \mathbb{N}$, the decision-maker may select a subset of basic actions to take (or \emph{arms to pull}) and receives a \emph{reward} which is a (possibly randomised) function of realisations of the random variables associated with the selected basic actions. The decision-maker's aim is to maximise her cumulative reward over a given horizon. Chen et al. study a CMAB problem where the decision-maker receives \emph{semibandit feedback} on her actions, meaning she observes the overall reward but also all realisations of the random variables associated with the selected arms. Realisations of the random variables are identically distributed for a given arm and independent both across time and arms.

In our adaptive searching problem, electing to search a cell $k$ in a round $t$, i.e. setting $a_{kt} \neq 0$, is the analogue of pulling an arm $k$. The total number of events detected in a round is the analogue of reward. The fundamental, and non-trivial difference between our model and that of Chen et al. lies in the feedback mechanism. Our framework is more complex in two important regards. Firstly, we do not by default observe independent identically distributed (i.i.d.) realisations of the underlying random variable of interest $X_{kt}$ each time we elect to search a cell. We observe a \emph{filtered observation} $Y_{kt}$ whose distribution depends on the action $\mathbf{a}_t$ selected in that round. This introduces complex dependencies within the sequence of rewards meaning standard concentration results for independent observations do not apply. Secondly, because of the $U$ possibly heterogeneous searchers, we can have multiple ways of searching the same collection of cells. While this is implicitly permitted within the framework of Chen et al., it is not explicitly acknowledged nor to the best of our knowledge are any real problems with such a structure explored in related work . 

Our analytical challenge is to extend earlier work in order to meet these novel features. Specifically we will propose a UCB algorithm for both cases of our problem and derive upper bounds on the expected regret of these policies. UCB algorithms apply the principle of \emph{optimism in the face of uncertainty} to sequential decision problems. Such an algorithm calculates an \emph{index} for each action in each round which is the upper limit of a high probability confidence interval on the expected reward of that action and then selects the action with the highest index. In this way the algorithm will select actions which either have high indices due to a large mean estimate - leading it to exploit what has been profitable so far - or due to a large uncertainty in the empirical mean - leading it to explore actions which are currently poorly understood. As the rounds proceed, the confidence intervals will concentrate on the true means and fewer exploratory actions will be selected in favour of exploitative ones.

\subsection{Case (I): Known detection probabilities} \label{sec::known_gammas}
In our first version of the problem, case (I), the only unknowns are the underlying rate parameters $\bs\lambda$. We assume that detection probability vectors $\bs\gamma(\mathbf{a})$ are known for all $\mathbf{a} \in \mathcal{A}$. Therefore we do not need to explicitly form UCB indices for every action separately. It will suffice to form a UCB index on each unknown $\lambda_k$ for $k \in [K]$. Optimistic estimates of the value of each action will then arise by calculating the $q_{\bs\lambda,\bs\gamma,iju}$ quantities with the optimistic estimate of $\bs\lambda$ in place of known $\bs\lambda$.

Our proposed approach to the sequential search problem in case (I), the FP-CUCB algorithm (Filtered Poisson - Combinatorial Upper Confidence Bound), is given as Algorithm \ref{alg::FPCUCB}. The algorithm consists of an initialisation phase of length $K$ where allocations are selected such that every cell is searched in some capacity at least once. Then in every subsequent round $t>K$, a UCB index \begin{equation}
\bar{\lambda}_{k,t}= \frac{\sum_{s=1}^{t-1}Y_{k,s}}{\sum_{s=1}^{t-1}\gamma_{k,s}}+\frac{6\max(1,\sqrt{\lambda_{max}})\log(t)}{\sum_{s=1}^{t-1}\gamma_{k,s}}+\sqrt{\frac{6\lambda_{max}\log(t)}{\sum_{s=1}^{t-1}\gamma_{k,s}}}, \label{eq::FPCUCBI_indices}
\end{equation} is calculated for each cell $k$. \CH{This particular UCB index is chosen because it can be shown to bound $\lambda_k$ with high probability. Specifically, using de la Pe\~{n}a's inequality \citep{DeLaPena1999}, it can be shown that $P(\bar\lambda_{k,t} \geq \lambda_k)$ approaches 1 as $t \rightarrow \infty$ at an appropriate rate.} A full derivation of this term is given in the proof of the following theorem.

An action which is optimal with respect to the $K$-vector of inflated rates $\bar{\bs\lambda}_t=(\bar{\lambda}_{1,t},...,\bar{\lambda}_{K,t})$ is then selected by solving the IP (\ref{eq::IP}) with $\bar{\bs\lambda}_t$ in place of $\bs\lambda$. The inflation terms involve a parameter $\lambda_{max} \geq \max_{k \in [K]} \lambda_k$. This is necessary to construct UCBs which concentrate at a rate that matches the concentration of Poisson random variables, which is defined by the mean parameter.

\begin{algorithm}[]
	\vspace{0.1cm}
	\caption{FP-CUCB (case (I))}
	\label{alg::FPCUCB}
	\vspace{0.1cm}
	\vspace{0.1cm}
	\textbf{Inputs:} Upper bound $\lambda_{max} \geq \lambda_k, \enspace k \in [K]$. 
	
	\textbf{Initialisation Phase:} For $t \in [K]$ 
	\begin{itemize}
	\item Select an arbitrary allocation $\mathbf{a} \in \mathcal{A}$ such that $a_t \neq 0$
	\end{itemize}
	
	\textbf{Iterative Phase:} For $t=K+1,K+2,...$
	\begin{itemize}
		\item Calculate indices \begin{equation*}
		\bar{\lambda}_{k,t}= \frac{\sum_{s=1}^{t-1}Y_{k,s}}{\sum_{s=1}^{t-1}\gamma_{k,s}}+\frac{6\max(1,\sqrt{\lambda_{max}})\log(t)}{\sum_{s=1}^{t-1}\gamma_{k,s}}+\sqrt{\frac{6\lambda_{max}\log(t)}{\sum_{s=1}^{t-1}\gamma_{k,s}}}, \enspace k \in [K] 
		\end{equation*}
		\item Select an allocation $\mathbf{a}^*_{\bar{\bs\lambda}_t}$ such that $r_{\bar{\bs\lambda}_t,\bs\gamma}(\mathbf{a}^*_{\bar{\bs\lambda}_t})=\max_{\mathbf{a} \in \mathcal{A}} r_{\bar{\bs\lambda}_t,\bs\gamma}(\mathbf{a})$. 
	\end{itemize}
\end{algorithm}

To analyse the regret of this algorithm we must first introduce some additional notation for \emph{optimality gaps}, the differences in expected reward between optimal and suboptimal actions. For $k \in [K]$ define, \begin{align*}
\Delta_{max}^k&=\text{opt}_{\bs\lambda,\bs\gamma}-\min_{\mathbf{a} \in \mathcal{A}_k}r_{\bs\lambda,\bs\gamma}(\mathbf{a}),  \\
\Delta_{min}^k&=\text{opt}_{\bs\lambda,\bs\gamma}-\max_{\mathbf{a} \in \mathcal{A}_k}r_{\bs\lambda,\bs\gamma}(\mathbf{a}),
\end{align*} where $\mathcal{A}_k=\{\mathbf{a} \in \mathcal{A}: a_k \neq 0\}$ for $k \in [K]$, and $\Delta_{max}=\max_{k \in [K]}\Delta^k_{max}$, and $\Delta_{min}=\min_{k \in [K]} \Delta^k_{min}$. The quantity $\Delta_{max}$ is then the difference in expected reward between an optimal allocation of searchers and the worst possible allocation, while $\Delta_{min}$ is the difference in expected reward between an optimal allocation and the closest to optimal suboptimal allocation. The quantities $\Delta^k_{max}$ and $\Delta^k_{min}$ are the largest and smallest gaps between the expected reward of an optimal allocation and allocations where cell $k$ is searched in some capacity. All $\Delta$ terms depend on $\bs\lambda, \bs\gamma$ but we drop this dependence in the notation for simplicity. 

\subsubsection{Upper bound on regret}

Now, in Theorem \ref{thm::FPCUCB} we provide an analytical bound on the expected regret of the {FP-CUCB} algorithm in $n$ rounds.

\begin{theorem} \label{thm::FPCUCB}
	The regret of the FP-CUCB algorithm with $\lambda_{max}$ applied to the sequential surveillance problem with known $\bs\gamma$ satisfies \begin{align}
	Reg_{\bs\lambda,\bs\gamma}^{\text{FP-CUCB}}(n) &\leq \sum_{k:\Delta_{min}^k>0}\frac{12K^2}{\gamma_{k,min}}  \Bigg[ \frac{b(\Delta^k_{min})}{\Delta_{min}^k}  + \int_{\Delta_{min}^k}^{\Delta_{max}^k}  \frac{b(x)}{x^2} dx \Bigg]\log(n) + \bigg(\frac{\pi^2}{3}+1\bigg)K\Delta_{max}, \label{eq::FPCUCBBound}
	\end{align}
where \begin{equation*}
b(x)=\lambda_{max} + \frac{x\max(1,\sqrt{\lambda_{max}})}{K} + \sqrt{\lambda_{max}^2 + \frac{2x\lambda_{max}\max(1,\sqrt{\lambda_{max}})}{K}},
\end{equation*} and $\gamma_{k,min}=\min_{\mathbf{a}:a_k \neq 0} \gamma_k(\mathbf{a})$.
\end{theorem}

To give a proof of this theorem we must introduce a new way of thinking about the action space. Consider that while we have previously (for ease of exposition) defined actions in terms of  allocations of searchers to cells, $\mathbf{a} \in \mathcal{A}$, the real impact on reward comes from the vectors of detection probabilities, $\bs\gamma(\mathbf{a})$, which arise from these allocations. As multiple allocations may give rise to the same vector of detection probabilities (if, for instance, two searchers have identical capabilities, then switching their assignments would have no impact on the quality of the search) the set $\mathcal{G} = \{\bs\gamma(\mathbf{a}), \enspace \forall \mathbf{a} \in \mathcal{A}\}$ of possible detection probability vectors most parsimoniously describes the set of possible actions in this problem. 

For an element \CH{$\mathbf{g}=(g_1,...,g_K)\in\mathcal{G}$} we then have expected reward $\mathbf{g}^T\cdot \bs\lambda$ and optimality gap $\Delta_{\mathbf{g}}=\text{opt}_{\bs\lambda,\bs\gamma}-\mathbf{g}^T \cdot \bs\lambda$. Let $\mathcal{G}_k$ be the set of vectors $\mathbf{g}$ with $g_k>0$ and $\mathcal{G}_{k,B}\subseteq\mathcal{G}_k$ be the set of vectors in $\mathcal{G}_k$ with sub-optimal expected reward - i.e. $\mathcal{G}_{k,B}=\{\mathbf{g} \in \mathcal{G}_k: \Delta_{\mathbf{g}}>0\}$. Let $B_k=|\mathcal{G}_{k,B}|$ and label the vectors in $\mathcal{G}_{k,B}$ as $\mathbf{g}^1_{k,B},\mathbf{g}^2_{k,B},...,\mathbf{g}^{B_k}_{k,B}$ in increasing order of expected reward. We use the following notation for optimality gaps with respect to these ordered vectors \begin{equation}
\Delta^{k,j} = \text{opt}_{\bs\lambda,\bs\gamma} - (\mathbf{g}^j_{k,B})^T \cdot \bs\lambda \quad j \in [B_k], \enspace k \in [K]
\end{equation}
and thus the gaps defined previously can be expressed as $\Delta_{max}^k = \Delta^{k,1}$ and $\Delta_{min}^k = \Delta^{k,B_k}$. We introduce counters $D_{k,t}=\sum_{s=1}^t g_{k,s}$ for $k \in [K]$, $t \in \mathbb{N}$ where $\mathbf{g}_s$ is the detection probability vector selected in round $s$. These allow us to keep track of the total detection probability \emph{applied} to a cell up to the end of round $t$.

The central idea in proving Theorem \ref{thm::FPCUCB} is that if for a certain sub-optimal action $\mathbf{g}: \Delta_{\mathbf{g}}>0$, all the cells $k$ with $g_k>0$ have been sampled sufficiently, the mean estimates ought to be accurate enough that the probability of selecting that sub-optimal action again before horizon $n$ is small. We show that this sufficient sampling level is $O(\log(n))$ and the ``small" probabilities of selecting the sub-optimal action after sufficient sampling are so small as to converge to a constant. Thus by re-expressing expected regret as a function of the number of plays of sub-optimal actions, we can bound it from above as the sum of a $O(\log(n))$ term derived from the sufficient sampling level and a constant independent of $n$. 

To count the plays of sub-optimal actions we maintain counters $N_{k,t}$, which collectively count the number of suboptimal plays. We update them as follows. Firstly, after the $K$ initialisation rounds we set $N_{k,K}=1$ for $k \in [K]$. Thereafter, in each round $t >K$, let $k'=\arg\min_{j:g_{j,t}>0}N_{j,t-1}$ \CH{(i.e. $k'$ indexes the cell involved in the current action which has the lowest counter)}, where if $k'$ is non-unique, we choose a single value randomly from the minimising set. If $\mathbf{g}_t^T\cdot \bs\lambda \neq \text{opt}_{\bs\lambda,\bs\gamma}$ then we increment $N_{k'}$ by one, i.e. set $N_{k',t}=N_{k',t-1}+1$. The key consequences of these updating rules are that $\sum_{k=1}^K N_{k,t}$ provides an upper bound on the number of suboptimal plays in $t$ rounds \CH{(since one of the first $K$ actions may be optimal)}, and $D_{k,t} \geq \gamma_{k,min}N_{k,t}$ for all $k$ and $t$ \CH{(since cell $k$ is always searched with detection probability at least $\gamma_{k,min}$)}. \CH{While tracking the sub-optimal plays in this way is more complex than maintaining a single counter of the number of sub-optimal actions, it permits a convenient decomposition of regret that allows us to prove Theorem \ref{thm::FPCUCB}.} \\

\emph{Proof of Theorem \ref{thm::FPCUCB}:} We prove the theorem by decomposing regret into a function of the number of plays of suboptimal arms, up to and after some sufficient sampling level. We then introduce two propositions which give bounds for quantities in the decomposition which are then combined to give the bound in (\ref{eq::FPCUCBBound}). The proofs of these propositions is reserved for Appendix \ref{Theorem1Proof}.

Let $N_{k,t}^{l,suf}, N_{k,t}^{l,und}$ for $l \in [B_k]$ be counters associated with elements of $\mathcal{G}_{k,B}$ for $k \in [K]$. These counters are defined as follows: \begin{align}
N_{k,n}^{l,suf} &= \sum_{t=K+1}^n \mathds{I}\{\mathbf{g}_t = \mathbf{g}_{k,B}^l, N_{k,t} > N_{k,t-1},N_{k,t-1}> h_{k,n}(\Delta^{k,l})\}, \label{Beyond Sufficiency Counter} \\
N_{k,n}^{l,und} &= \sum_{t=K+1}^n \mathds{I}\{\mathbf{g}_t = \mathbf{g}_{k,B}^l, N_{k,t} > N_{k,t-1},N_{k,t-1}\leq  h_{k,n}(\Delta^{k,l})\}, \label{Under Sufficiency Counter}
\end{align}
where $h_{k,n}(\Delta)=12b(\Delta) \frac{\log(n)K^2}{\gamma_{k,min}\Delta^2}$. A cell $k$ is said to be \emph{sufficiently} sampled with respect to a choice of detection probabilities $\mathbf{g}_{k,B}^l$ if $N_{k,t-1} > h_{k,n}(\Delta^{k,l})$, and thus $N_{k,n}^{l,und},N_{k,n}^{l,suf}$ count the suboptimal plays leading to incrementing $N_{k,n}^l$ up to and after the sufficient level, respectively. 

From the definitions (\ref{Beyond Sufficiency Counter}) and (\ref{Under Sufficiency Counter}) we have $N_{k,n}=1+\sum_{l=1}^{B_k} (N_{k,n}^{l,suf}+N_{k,n}^{und})$. The expected regret at time horizon $n$ can also be bounded above using this notation as \begin{equation}
Reg_{\bs\lambda,\bs\gamma}(n) \leq \mathds{E}\Bigg[\sum_{k=1}^K \Bigg(\Delta^{k,1} + \sum_{l=1}^{B_k}(N_{k,n}^{l,suf}+N_{k,n}^{l,und})\cdot \Delta^{k,l}\Bigg)\Bigg] \label{eq::regret_decomposition}
\end{equation}
where $\Delta^{k,1}$ arises as a worst case view of the initialisation. We can derive an analytical bound on regret by bounding the expectations of the random variables in (\ref{eq::regret_decomposition}). 

Firstly, for the beyond sufficiency counter we have 
\begin{proposition} \label{prop::FPCUCBsuff}
	For any time horizon $n>K$, \begin{equation}
	\mathds{E}\Bigg(\sum_{k=1}^K \sum_{l=1}^{B_k} N_{k,n}^{l,suf}\Bigg) \leq \frac{\pi^2}{3}\cdot K. \label{eq::prop1}
	\end{equation} 
\end{proposition}
The full proof of Proposition \ref{prop::FPCUCBsuff} is given in Appendix \ref{Theorem1Proof}, but it depends in particular on the following Lemma describing the concentration of filtered Poisson data. The derivation of the concentration result for the observations $Y_1,...,Y_t$ requires careful treatment as the parameters of these distributions, and therefore the observations themselves, are not independent. The stochastic dependencies between the sequence of random variables $\gamma_1,...,\gamma_s$ may be highly complex, so rather than attempt to quantify these relationships exactly, we appeal to martingale theory which allows us to derive the concentration result without assuming independence. We provide the necessary concentration result in the lemma below.

\begin{lemma} \label{lem::FPCUCBconc}
	Let $Y_{1},...,Y_{s}$ be any sequence of Poisson random variables with means $\gamma_{1}\lambda,...\gamma_{s}\lambda$ respectively, such that the sequence $\{Z_j\}_{j=1}^s$ is a martingale where $Z_j=\sum_{i=1}^j (Y_i - \mathbb{E}(Y_i|Y_{i-1},...,Y_1))$. Then, given parameters $t\geq s$ and $\lambda_{max} \geq \lambda$ the following holds:
	\begin{equation}
	\mathds{P}\bigg(\bigg|\frac{\sum_{j=1}^{s}Y_{j}}{\sum_{j=1}^s\gamma_{j}}-\lambda\bigg| \geq \frac{6\max(1,\sqrt{\lambda_{max}})\log(t)}{\sum_{j=1}^s\gamma_{j}}+\sqrt{\frac{6\lambda_{max}\log(t)}{\sum_{j=1}^s\gamma_{j}}}\bigg) \leq 2t^{-3}. \label{FPCUCBconc}
	\end{equation}
\end{lemma}

The proof of this Lemma is given in Appendix \ref{proof::concresult}. The consequence of this Lemma is that the UCB indices (\ref{eq::FPCUCBI_indices}) are of the correct form to guarantee that the probability of making suboptimal plays beyond the sufficient sampling level is small. 

For the under sufficiency counter we have the following proposition, also proved in Appendix \ref{Theorem1Proof},
\begin{proposition} \label{prop::FPCUCBund}
	For any time horizon $n>K$ and $k:\Delta^k_{min}>0$, \begin{equation}
	\sum_{l=1}^{B_k} N_{k,n}^{l,und}\Delta^{k,l} \leq h_{k,n}(\Delta^{k,B_k})\Delta^{k,B_k} + \int_{\Delta^{k,B_k}}^{\Delta^{k,1}} h_{k,n}(x)dx. \label{eq::prop2}
	\end{equation} 
\end{proposition}
Combining the decomposition (\ref{eq::regret_decomposition}), with the bounds (\ref{eq::prop1}) and (\ref{eq::prop2}) we have \begin{align*}
Reg_{\bs\lambda,\bs\gamma}(n) &\leq  \mathds{E}\bigg(\sum_{k=1}^K \Big(\Delta^{k,1}+\sum_{l=1}^{B_k}(N_{k,n}^{l,suf}+N_{k,n}^{l,und})\Delta^{k,l}\Big)\bigg) \\
&= \mathds{E}\Bigg(\sum_{k=1}^K\bigg(\Delta^{k,1} + \sum_{l=1}^{B_k} N_{k,n}^{l,suf}\Delta^{k,l}\bigg)\Bigg) + \mathds{E}\Bigg(\sum_{k=1}^K \sum_{l=1}^{B_k} N_{k,n}^{l,und}\Delta^{k,l}\Bigg) \\
&\leq K\Delta_{max} + \mathds{E}\Bigg(\sum_{k=1}^K \sum_{l=1}^{B_k} N_{k,n}^{l,suf}\Delta^{k,l}\Bigg) + \sum_{k:\Delta_{min}^k>0} \bigg(h_{k,n}(\Delta^{k,B_k})\Delta^{k,B_k} + \int_{\Delta^{k,B_k}}^{\Delta^{k,1}} h_{k,n}(x)dx\bigg) \\
&\leq \Big(\frac{\pi^2}{3}+1\Big)K \Delta_{max} + \sum_{k:\Delta_{min}^k>0} \bigg(h_{k,n}(\Delta^{k}_{min})\Delta^{k}_{min} + \int_{\Delta^{k}_{min}}^{\Delta^{k}_{max}} h_{k,n}(x)dx\bigg) \\
&=\sum_{k:\Delta_{min}^k>0}\frac{12K^2}{\gamma_{k,min}}  \Bigg[ \frac{b(\Delta^k_{min})}{\Delta_{min}^k}  + \int_{\Delta_{min}^k}^{\Delta_{max}^k}  \frac{b(x)}{x^2} dx \Bigg]\log(n) + \bigg(\frac{\pi^2}{3}+1\bigg)K\Delta_{max}. \quad \square
\end{align*}

In the remainder of this section we show that the bound obtained in Theorem \ref{thm::FPCUCB} is of optimal order, by deriving a lower bound on the expected regret of the best possible policies. We also proceed to show a second upper bound of sub-optimal order with respect to $n$ but that has the advantage of holding for any problem instance, and therefore does not depend on the optimality gaps, $\Delta_{min}^k$ and $\Delta_{max}^k$, $\forall k \in [K]$.

\subsubsection{Lower Bound on Regret} \label{sec::regret_lower_bound}
To analyse the performance of the best possible policies, we introduce the notion of a \emph{uniformly good policy}. A uniformly good policy \citep{Lai1985asymptotically} is one where \begin{displaymath}
\mathds{E}\bigg(\sum_{t=1}^n \mathds{I}\{\mathbf{g}_t=\mathbf{g} \} \bigg) = o(n^\alpha) \quad \forall \enspace \alpha >0 
\end{displaymath} for every $\mathbf{g}: \Delta_{\mathbf{g}}>0$ and every $\bs\lambda \in \mathbb{R}_+^K$. Clearly, then all uniformly good policies must eventually favour optimal actions over suboptimal ones - with the suboptimal actions being necessary to accurately estimate $\bs\lambda$.  For a given rate vector $\bs\lambda$ we define the set of optimal actions as \begin{displaymath}
J(\bs\lambda)=\{\mathbf{g}\in\mathcal{G}:\enspace \mathbf{g}^T\cdot\bs\lambda=\text{opt}_{\bs\lambda,\bs\gamma}\}.
\end{displaymath} We write $S(\bs\lambda)=\mathcal{G}\setminus J(\bs\lambda)$ to be the set of suboptimal actions.  The difficulty of a particular problem depends on the particular configuration of $\bs\lambda$ and $\bs\gamma$. We define \begin{displaymath}
\mathcal{I}(\bs\lambda)= \{k: \enspace \exists \text{ } \mathbf{g} \in J(\bs\lambda) \text{ s.t. } g_k >0\}
\end{displaymath}
as the set of arms which are played in at least one optimal action and
 \begin{displaymath}
B(\bs\lambda)=\{\bs\theta \in \mathbb{R}_+^K : \enspace  \mathbf{g}^T \cdot \bs\theta  < \text{opt}_{\bs\theta,\bs\gamma} \enspace \forall  \mathbf{g} \in J(\bs\lambda) \enspace \text{ and } \enspace \theta_k = \lambda_k \enspace \forall  k \in \mathcal{I}(\bs\lambda)\}
\end{displaymath}
as the set of mean vectors such that all actions in $J(\bs\lambda)$ are suboptimal but this cannot be discerned by playing only actions in $J(\bs\lambda)$. The larger the set $B(\bs\lambda)$, the more challenging the problem is. If $\mathcal{I}(\bs\lambda)=[K]$, then the problem is trivial as one can simultaneously play optimal actions and gather the information necessary to affirm that these actions are optimal. In such a case the lower bound on expected regret is simply 0.

We have the following lower bound on regret for any uniformly good policy. A key consequence of this result is the assertion that policies with $O(\log(n))$ regret are indeed of optimal order and thus that the regret induced by the FP-CUCB algorithm in case (I) grows at the lowest achievable rate. This result is analogous to results in other classes of bandit problem as shown by \cite{Lai1985asymptotically} and \cite{Burnetas1996optimal}.

\begin{theorem} \label{thm::lower_bound}
For any $\bs\lambda \in \mathds{R}_+^K$ such that $B(\bs\lambda)\neq \emptyset$, and for any uniformly good policy $\pi$ for the sequential surveillance problem with known $\bs\gamma$, we have \begin{equation}
\lim \inf_{n \rightarrow \infty} \frac{Reg^\pi_{\bs\lambda,\bs\gamma}(n)}{\log(n)} \geq c(\bs\lambda)
\end{equation} 
where $c(\bs\lambda)$ is the optimal value of the following optimisation problem \CH{over non-negative coefficients $\mathbf{d}=\{d_\mathbf{g}, \mathbf{g} \in S(\bs\lambda)\}$,} \begin{align}
 \inf_{\mathbf{d}\geq \mathbf{0}} \sum_{\mathbf{g} \in S(\bs\lambda)} d_{\mathbf{g}}\Delta_{\mathbf{g}} & \label{eq::lb_opt_objective}\\
\text{such that } \inf_{\bs\theta \in B(\bs\lambda)} \sum_{\mathbf{g} \in S(\bs\lambda)} d_{\mathbf{g}} \sum_{k=1}^K g_k kl(\lambda_k,\theta_k) &\geq 1. \label{eq::lb_opt_constraint}
\end{align}
and $\text{kl}(\lambda,\theta)=\lambda\log(\frac{\lambda}{\theta})+\theta-\lambda$ is the Kullback Leibler divergence between two Poisson distributions with mean parameters $\lambda$, $\theta$ respectively.
\end{theorem}

We prove this theorem fully in Appendix \ref{proof::lower_bound}, but here note that a key step of its proof is to invoke Theorem 1 of \cite{Graves1997}, which is a similar result for a more general class of controlled Markov Chains. It is possible to derive an analytical expression giving a lower bound on $c(\bs\lambda)$ by following steps similar to those in the proof of Theorem 2 in \cite{Combes2015}. However we omit this here in the interests of succinctness as it is not an especially useful or elegant expression. 

\CH{We note that the lower bound is based on the KL-divergence of the cell means, and this suggests that, as in simpler MAB problems, an approach incorporating the KL-divergence in the UCB indices could be asymptotically optimal. However, existing theory on the convergence of such approaches \citep{Garivier2011,Combes2015} pertains only to the case of independent reward generation action selection mechanisms. Therefore, it is not clear how to approach the optimal design and analysis of such an approach.}

\subsubsection{Gap-free bound on regret}
The logarithmic order bounds \CH{of Theorems \ref{thm::FPCUCB} and \ref{thm::lower_bound}} are useful as they establish the order-optimality of the UCB algorithm. \CH{We note that the coefficients of the two bounds are not the same, and the upper bound may be very large in problem instances where the $\Delta_{min}^k$ terms are very small. 

The main purpose, however, of the bounds in Theorems \ref{thm::FPCUCB} and \ref{thm::lower_bound} is analytical, not numerical as they can be challenging to compute in practice. The computation of the $\Delta_{min}^k$ and $\Delta_{max}^k$ terms used in the upper bound requires evaluating the expected reward of every possible action, which quickly becomes computationally challenging for even modest values of $K$ and $U$. The computation of the lower bound again requires computation of the expected reward of every possible action, to calculate the $\Delta_{\mathbf{g}}$ terms, and also a minimisation over $|S(\bs\lambda)|$ variables, subject to a non-linear constraint. This optimisation problem lacks an convenient analytical solution and must be resolved numerically. 

Moreover, in absence of knowledge of the true reward generating parameters these bounds do little to inform one of expected performance of the algorithm.} For these reasons, we also present the following upper bound on regret, which is order-suboptimal, being of order $O(K\sqrt{n\log(n)})$, but holds uniformly across any choice of $\bs\lambda \in [0,\lambda_{max}]^K$ and does not depend on the optimality gaps. 

\begin{theorem} \label{thm::gapfree}
	The regret of the FP-CUCB algorithm with $\lambda_{max}$ applied to the sequential surveillance problem with known $\bs\gamma$ satisfies \begin{align}
	Reg_{\bs\lambda,\bs\gamma}^{FP-CUCB}(n) &\leq \frac{5K\lambda_{max}}{2}+ \frac{12K\max(1,\sqrt{\lambda_{max}})}{\gamma_{min}}\log(n)(1+\log(n)) \nonumber \\
	&\quad \quad +\sqrt{\frac{92K^2\lambda_{max}n\log(n)}{\gamma_{min}}}.
	\end{align}
\end{theorem} 

\noindent
\emph{Proof of Theorem \ref{thm::gapfree}:} We first consider the following decomposition of regret,
\begin{align}
Reg_{\bs\lambda,\bs\gamma}^{FP-CUCB}(n) &= \mathbb{E}\bigg(\sum_{t=1}^n r_{\bs\lambda,\bs\gamma}(\mathbf{a}^*)-r_{\bs\lambda,\bs\gamma}(\mathbf{a}_t)\bigg)\nonumber \\
&= \mathbb{E}\bigg(\sum_{t=1}^n r_{\bs\lambda,\bs\gamma}(\mathbf{a}^*)-r_{\bs\lambda,\bs\gamma}(\mathbf{a}_t) + r_{\bar{\bs\lambda}_t,\bs\gamma}(a_t)- r_{\bar{\bs\lambda}_t,\bs\gamma}(a_t) \bigg)\nonumber \\
&= \mathbb{E}\bigg(\sum_{t=1}^n \sum_{k=1}^K \lambda_k g_k^* - \bar\lambda_{kt}g_{kt}+\bar\lambda_{kt}g_{kt} - \lambda_k g_{kt}\bigg)\nonumber \\
&\leq \mathbb{E}\bigg(\sum_{t=1}^n \sum_{k=1}^K (\lambda_k-\bar\lambda_{kt}) g_k^* +(\bar\lambda_{kt}-\lambda_k)g_{kt}\bigg) \nonumber \\
&= \sum_{t=1}^n \sum_{k=1}^K\mathbb{E}\bigg( (\lambda_k-\bar\lambda_{kt})\bigg)g_k^* + \sum_{t=1}^n \sum_{k=1}^K\mathbb{E}\bigg((\bar\lambda_{kt}-\lambda_k)g_{kt}\bigg) \label{eq::gapfreedecomp}
\end{align}
The terms of the first sum in \eqref{eq::gapfreedecomp} are very unlikely to be positive, increasingly so as more data is collected. If we upper bound by ignoring the case of negative terms we have: \begin{align*}
&\enspace \sum_{t=1}^n \sum_{k=1}^K\mathbb{E}\bigg( (\lambda_k-\bar\lambda_{kt})\bigg)g_k^* \\
&\leq \sum_{t=1}^n \sum_{k=1}^K g_k^* \mathbb{P}(\lambda_k > \bar\lambda_{kt})\mathbb{E}(\lambda_k - \bar\lambda_{kt}| \lambda_k > \bar\lambda_{kt}) \\ 
&\leq \sum_{t=1}^n \lambda_{max} \sum_{k=1}^K \mathbb{P}(\lambda_k > \bar\lambda_{kt}) \\
&= K\lambda_{max} \sum_{t=1}^n  \mathbb{P}\bigg( \sum_{k=1}^K \lambda_k > \sum_{k=1}^K \frac{\sum_{s=1}^{t-1}Y_{k,s}}{\sum_{s=1}^{t-1}\gamma_{k,s}}+ \frac{6\max(1,\sqrt{\lambda_{max}})\log(t)}{\sum_{s=1}^{t-1}\gamma_{k,s}}+ \sqrt{\frac{6\lambda_{max}\log(t)}{\sum_{s=1}^{t-1} \gamma_{k,s}}}\bigg) \\
&\leq K\lambda_{max} \sum_{t=1}^n t^{-3} \leq \frac{5K\lambda_{max}}{4}
\end{align*} where the penultimate inequality is due to Lemma \ref{lem::FPCUCBconc}.

Now consider the second sum in \eqref{eq::gapfreedecomp} \begin{align*}
&\quad \sum_{t=1}^n \sum_{k=1}^K\mathbb{E}\bigg((\bar\lambda_{kt}-\lambda_k)g_{kt}\bigg) \\
&= \sum_{t=1}^n \mathbb{E}\bigg(\sum_{k=1}^K \gamma_{k,t} \Big( \frac{\sum_{s=1}^{t-1}Y_{k,s}}{\sum_{s=1}^{t-1}\gamma_{k,s}}+ \frac{6\max(1,\sqrt{\lambda_{max}})\log(t)}{\sum_{s=1}^{t-1}\gamma_{k,s}}+ \sqrt{\frac{6\lambda_{max}\log(t)}{\sum_{s=1}^{t-1} \gamma_{k,s}}} - \lambda_k\Big)\bigg) \\
&\leq \sum_{t=1}^n K\lambda_{max} \mathbb{P}\bigg(\sum_{k=1}^K \frac{\sum_{s=1}^{t-1}Y_{k,s}}{\sum_{s=1}^{t-1}\gamma_{k,s}}-\lambda_k > \sum_{k=1}^K \frac{6\max(1,\sqrt{\lambda_{max}})\log(t)}{\sum_{s=1}^{t-1}\gamma_{k,s}}+ \sqrt{\frac{6\lambda_{max}\log(t)}{\sum_{s=1}^{t-1} \gamma_{k,s}}} \bigg) \\
&\quad \quad + \sum_{t=1}^n \mathbb{E}\bigg(\sum_{k=1}^K 2\gamma_{k,t}\Big(\frac{6\max(1,\sqrt{\lambda_{max}})\log(t)}{\sum_{s=1}^{t-1}\gamma_{k,s}}+ \sqrt{\frac{6\lambda_{max}\log(t)}{\sum_{s=1}^{t-1} \gamma_{k,s}}}\Big)\bigg) \\
&\leq \frac{5K\lambda_{max}}{4}+ \mathbb{E}\bigg( \sum_{t=1}^n\sum_{k=1}^K \frac{12\gamma_{k,t}\max(1,\sqrt{\lambda_{max}})\log(t)}{\sum_{s=1}^{t-1}\gamma_{k,s}}\bigg)+ \mathbb{E}\bigg( \sum_{t=1}^n \sum_{k=1}^K \gamma_{k,t}\sqrt{\frac{24\lambda_{max}\log(t)}{\sum_{s=1}^{t-1} \gamma_{k,s}}}\bigg) \\
&\leq \frac{5K\lambda_{max}}{4}+ 12\max(1,\sqrt{\lambda_{max}})\log(n)\mathbb{E}\bigg(\sum_{t=1}^n \sum_{k=1}^K \frac{\gamma_{k,t}}{\sum_{s=1}^{t-1}\gamma_{k,s}}\bigg) \\
&\quad \quad + \sqrt{24\lambda_{max}\log(n)}\mathbb{E}\bigg(\sum_{t=1}^n\sum_{k=1}^K \frac{\gamma_{k,t}}{\sqrt{\sum_{s=1}^{t-1}\gamma_{k,s}}}\bigg)
\end{align*}
Consider the expectation in the final term, we have, \begin{align*}
\mathbb{E}\bigg(\sum_{t=1}^n\sum_{k=1}^K \frac{\gamma_{k,t}}{\sqrt{\sum_{s=1}^{t-1}\gamma_{k,s}}}\bigg) \leq \sum_{k=1}^K \sum_{t=2}^n \frac{1}{\sqrt{(t-1)\gamma_{min}}} \leq \frac{2K}{\sqrt{\gamma_{min}}}\sqrt{n}.
\end{align*} Similarly for the other expectation, we have \begin{align*}
\mathbb{E}\bigg(\sum_{t=1}^n \sum_{k=1}^K \frac{\gamma_{k,t}}{\sum_{s=1}^{t-1}\gamma_{k,s}}\bigg) \leq \sum_{k=1}^K \sum_{t=2}^n \frac{1}{(t-1)\gamma_{min}} \leq \frac{K}{\gamma_{min}}(1+\log(n)).
\end{align*}
Pulling this all together we have the following gap-free bound on regret: \begin{align*}
Reg_{\bs\lambda,\bs\gamma}^{FP-CUCB}(n) &\leq \frac{5K\lambda_{max}}{2} + \frac{12K\max(1,\sqrt{\lambda_{max}})}{\gamma_{min}}\log(n)(1+\log(n))\\
&\quad \quad  +\sqrt{\frac{92K^2\lambda_{max}\log(n)n}{\gamma_{min}}},
\end{align*} as stated in Theorem \ref{thm::gapfree}. $\square$

\subsection{Case (II): Known scaling of detection probabilities} \label{sec::unknown_gammas}

In the second case we suppose that we do not know exactly what probability of successful detection each searcher has in each cell, but that we have some idea of how these detection probabilities change as the searchers are assigned more cells to search. If, for example, the searcher is moving back-and-forth over $l$ cells at a constant speed $s$, then the time between successive visits to a cell is $2l/s$, suggesting that the detection probability may decay like $s/(2l)$ with the number of cells $l$.

In order to be precise about this case we suppose that detection probabilities have the form \begin{equation}
\gamma_k(\mathbf{a}) = \sum_{u=1}^U \phi_{u}(\mathbf{a}) \omega_{ku}\mathds{I}\{a_k=u\}, \enspace k \in [K], \label{eq::case_b_detec_probs}
\end{equation}
where $\phi_{u}: \mathcal{A} \rightarrow [0,1]$ are known \emph{scaling functions}, and $\omega_{ku} \in (0,1] \enspace \forall k \in [K], u \in [U]$ are unknown \emph{baseline detection probabilities} - the probability of searcher $u$ detecting events in cell $k$ given that is the only cell they are assigned to search. Functions $\phi_{u}$ are assumed to be decreasing in the number of cells searcher $u$ must search. For instance, and as suggested in the preceding paragraph, one suitable function may be $\phi_u(\mathbf{a})=(\sum_{k=1}^K \mathds{I}\{a_k=u\})^{-1}$, the reciprocal of the number of cells the searcher $u$ is assigned. Searcher effectiveness may however decay more slowly as the number of cells assigned grows if for instance events are visible for an extended period of time.

In case (II) the action set and observed rewards remain entirely the same as for case (I), it is the information initially available to the controller that differs. Here, both $\bs\lambda$, the $K$-vector of rate parameters, and $\bs\omega=(\omega_{1,1},...,\omega_{1,U},\omega_{2,1}...,\omega_{K,U})$, the $KU$-vector of baseline detection probabilities are unknown as opposed to solely $\bs\lambda$ in case (I). Due to nonidentifiability we cannot make direct inference on $\bs\lambda$ or $\bs\omega$. However, simply estimating the products of certain components is sufficient for optimal decision making as estimating the expected reward does not depend on having separate estimates of each parameter. Therefore we can simply consider $KU$ unknowns $\bs\tau=(\omega_{1,1}\lambda_1,...,\omega_{1,U}\lambda_1,\omega_{2,1}\lambda_2,...,\omega_{K,U}\lambda_K)$ when referring to the unknown parameters. 

As such this second case of the sequential search problem can also be modelled as a CMAB problem with filtered feedback. The set of arms is given by searcher-cell pairs $ku \in [K]\times[U]$. Each arm $ku$ is associated with a Poisson distribution with unknown parameter $\tau_{ku}=\omega_{k,u}\lambda_k$. We continue to use $\mathcal{A}$ to specify the action set and filtering is governed by scaling function vectors $\bs\phi(\mathbf{a})=(\phi_1(\mathbf{a}),...,\phi_U(\mathbf{a}))$. Let $\phi_{ku,t}$ denote the filtering probability associated with the searcher-cell pair $ku$ in round $t$. It is 0 if $a_{k,t}\neq u$ and $\phi_u(\mathbf{a}_t)$ if $a_{k,t}=u$.

Let reward in this setting be defined \begin{displaymath}
r_{\bs\lambda,\bs\gamma}(\mathbf{a})=\tilde{r}_{\bs\tau,\bs\phi}(\mathbf{a})=\sum_{u=1}^U \phi_u(\mathbf{a}) \sum_{k=1}^K \tau_{ku}\mathds{I}\{a_k=u\}
\end{displaymath} and define optimality gaps in this setting for $ku \in [K] \times [U]$ as\begin{align*}
\Delta_{max}^{ku}&=\text{opt}_{\bs\lambda,\bs\gamma}-\min_{\mathbf{a} \in \mathcal{A}}\{r_{\bs\lambda,\bs\gamma}(\mathbf{a}) \enspace | \enspace r_{\bs\lambda,\bs\gamma}(\mathbf{a}) \neq \text{opt}_{\bs\lambda,\bs\gamma},a_k=u\} \\
\Delta_{min}^{ku}&=\text{opt}_{\bs\lambda,\bs\gamma}-\max_{\mathbf{a} \in \mathcal{A}}\{r_{\bs\lambda,\bs\gamma}(\mathbf{a}) \enspace | \enspace r_{\bs\lambda,\bs\gamma}(\mathbf{a}) \neq \text{opt}_{\bs\lambda,\bs\gamma},a_k=u\}. 
\end{align*}

The appropriate FP-CUCB algorithm for case (II) then calculates upper confidence bounds for each $\tau_{ku}$ parameter instead of $\lambda_k$ and as in the FP-CUCB algorithm for case (I) this induces an optimistic estimate of the value of every $\mathbf{a} \in \mathcal{A}$. We describe this second variant in Algorithm \ref{alg::FPCUCBII}.

\begin{algorithm}[]
	\vspace{0.1cm}
	\caption{FP-CUCB (case (II))}
		\label{alg::FPCUCBII}
	\vspace{0.1cm}
	\vspace{0.1cm}
	\textbf{Inputs:} Upper bound $\tau_{max} \geq \tau_{ku}, \enspace k \in [K]$ and $u \in [U]$. 
	
	\textbf{Initialisation Phase:} For $t \in [KU]$ 
	\begin{itemize}
	\item Select an arbitrary allocation $\mathbf{a} \in \mathcal{A}$ such that $a_t\neq 0$
	\end{itemize}
	
	\textbf{Iterative Phase:} For $t=KU+1,KU+2,...$
	\begin{itemize}
		\item Calculate indices \begin{displaymath}
		\bar{\tau}_{ku,t}= \frac{\sum_{s=1}^{t-1}Y_{ku,s}}{\sum_{s=1}^{t-1}\phi_{ku,s}}+\frac{6\max(1,\sqrt{\tau_{max}})\log(t)}{\sum_{s=1}^{t-1}\phi_{ku,s}}+\sqrt{\frac{6\tau_{max}\log(t)}{\sum_{s=1}^{t-1}\phi_{ku,s}}}, \enspace ku \in [K]\times[U]
		\end{displaymath}
		\item Select an allocation $\mathbf{a}^*_{\bar{\bs\lambda}_t}$ such that $\tilde{r}_{\bar{\bs\tau}_t,\bs\phi}(\mathbf{a}^*_{\bar{\bs\lambda}_t})=\max_{\mathbf{a} \in \mathcal{A}} \tilde{r}_{\bar{\bs\tau}_t,\bs\phi}(\mathbf{a})$. 
	\end{itemize}
\end{algorithm}

Since our CMAB model in case (II) and second variant of FP-CUCB are of the same form as in case (I), the analogous results to Theorems \ref{thm::FPCUCB} and \ref{thm::lower_bound} can be derived. Specifically we have a regret upper bound for FP-CUCB in Corollary \ref{cor::FPCUCBb} and a lower bound for regret of any uniformly good algorithm in Corollary \ref{cor::lowerbound}.

\begin{corollary} \label{cor::FPCUCBb}
	The regret of the FP-CUCB algorithm in case (b) defined by $\tau_{max}$ applied to the sequential search problem as defined previously satisfies \begin{align*}
	Reg_{\bs\lambda,\bs\gamma}^{\text{FP-CUCB}}(n) &\leq \sum_{ku:\Delta_{min}^{ku}>0}\frac{12(KU)^2}{\phi_{ku,min}}  \Bigg[ \frac{b'(\Delta^{ku}_{min})}{\Delta_{min}^{ku}}  + \int_{\Delta_{min}^{ku}}^{\Delta_{max}^{ku}}  \frac{b'(x)}{x^2} dx \Bigg]\log(n) + \bigg(\frac{\pi^2}{3}+1\bigg)KU\Delta_{max}, 
	\end{align*}
where \begin{equation*}
\tilde{b}(x)=\tau_{max} + \frac{x\max(1,\sqrt{\tau_{max}})}{KU} + \sqrt{\tau_{max}^2 + \frac{2x\tau_{max}\max(1,\sqrt{\tau_{max}})}{KU}},
\end{equation*} and $\phi_{ku,min}=\min_{\mathbf{a}:a_k=u}\phi_u(\mathbf{a})$.
\end{corollary}

\begin{corollary} \label{cor::lowerbound}
For any $\bs\tau \in \mathds{R}_+^{KU}$ such that $\tilde{B}(\bs\tau)\neq \emptyset$, and for any uniformly good policy $\pi$ for the sequential surveillance problem with known $\bs\phi$, we have \begin{displaymath}
\lim \inf_{n \rightarrow \infty} \frac{Reg^\pi_{\bs\lambda,\bs\gamma}(n)}{\log(n)} \geq \tilde{c}(\bs\tau)
\end{displaymath} 
where $\tilde{c}(\bs\tau)$ is the solution of an optimisation problem analogous to (\ref{eq::lb_opt_objective}).
\end{corollary}
Precise specification of $\tilde{c}(\bs\tau)$ requires redefining notation from Section \ref{sec::regret_lower_bound} in the context of case (II) and produces an entirely unsurprising analogue. In the interests of brevity we omit this. The techniques used in proving Theorems \ref{thm::FPCUCB} and \ref{thm::lower_bound} can be easily extended to prove Corollaries 1 and 2.

\section{Numerical Experiments} \label{sec::numerics}
We now numerically evaluate the performance of the FP-CUCB algorithm in comparison to a greedy approach and Thompson Sampling (TS). The greedy approach is one which always selects the action currently believed to be best (following an initialisation period, where each cell is searched at least once). As such it is a fully exploitative policy which fails to recognise the benefit of the information gain inherent in exploration. TS is a randomised, Bayesian approach where an action is selected with the current posterior probability that it is the best one. This is achieved by sampling indices from a posterior distribution on each arm and passing these samples to the optimisation algorithm. We define these algorithms in the setting of known detection probabilities (case (I)) in Algorithms \ref{alg::Greedy} and \ref{alg::TS} respectively.

\begin{algorithm}[]
	\vspace{0.1cm}
	\caption{Greedy}
	\label{alg::Greedy}
	\vspace{0.1cm}
	\vspace{0.1cm}
	\textbf{Initialisation Phase:} For $t \in [K]$ 
	\begin{itemize}
	\item Select an arbitrary allocation $\mathbf{a} \in \mathcal{A}$ such that $a_t \neq 0$
	\end{itemize}
	
	\textbf{Iterative Phase:} For $t=K+1,K+2,...$
	\begin{itemize}
		\item For each $k \in [K] $ calculate $\hat{\lambda}_{k,t}= \frac{\sum_{s=1}^{t-1}Y_{k,s}}{\sum_{s=1}^{t-1}\gamma_{k,s}}$
		\item Select an allocation $\mathbf{a}^*_{\hat{\bs\lambda}_t}$ such that $r_{\hat{\bs\lambda}_t,\bs\gamma}(\mathbf{a}^*_{\hat{\bs\lambda}_t})=\max_{\mathbf{a} \in \mathcal{A}} r_{\hat{\bs\lambda}_t,\bs\gamma}(\mathbf{a})$. 
	\end{itemize}
\end{algorithm}

\begin{algorithm}[]
	\vspace{0.1cm}
	\caption{Thompson Sampling (TS)}
	\label{alg::TS}
	\vspace{0.1cm}
	\vspace{0.1cm}
	\textbf{Inputs:} Gamma prior parameters $\alpha, \beta$
	
	\textbf{Iterative Phase:} For $t=1,2,...$
	\begin{itemize}
		\item For each $k \in [K]$ sample $\tilde{\lambda}_{k,t}$ from a $\text{Gamma}(\alpha + \sum_{s=1}^{t-1}Y_{k,s}, \beta + \sum_{s=1}^{t-1} \gamma_{k}(\mathbf{a}_s))$ distribution
		\item Select an allocation $\mathbf{a}^*_{\tilde{\bs\lambda}_t}$ such that $r_{\tilde{\bs\lambda}_t,\bs\gamma}(\mathbf{a}^*_{\tilde{\bs\lambda}_t})=\max_{\mathbf{a} \in \mathcal{A}} r_{\tilde{\bs\lambda}_t,\bs\gamma}(\mathbf{a})$. 
	\end{itemize}
\end{algorithm}

We compare the FP-CUCB, Greedy and TS algorithms by randomly sampling $\bs\lambda$ and $\bs\omega$ values which define problem instances. We then test our algorithms' performance on data generated from the models of these problem instances. We assume that detection probabilities have the form given in (\ref{eq::case_b_detec_probs}) but we know both the $\phi$ functions and $\omega$ values. 

Specifically, we conduct four tests encompassing a range of different problem sizes and parameter values to display the efficacy of our proposed approach uniformly across problem instances. In each test 50 $(\bs\lambda,\bs\omega)$ pairs are sampled and functions $\bs\phi$ are selected. For each $(\bs\lambda,\bs\omega)$ pair 5 datasets are sampled giving underlying counts of intrusion events in each cell in each round up to a horizon of $n=2000$. Parameters are simulated as below: \begin{enumerate}
\item[(i)] $K=15$ cells and $U=5$ searchers. Cell means $\lambda_k$ are sampled from a $\text{Uniform}(10,20)$ distribution for $k \in [K]$. Baseline detection probabilities $\omega_{ku}$ are sampled from $\text{Beta}(u,2)$ distributions for $u \in [U]$, $k \in [K]$. Scaling functions are $\phi_u(\mathbf{a})=(\sum_{k=1}^K \mathbb{I}\{a_k=u\})^{-1}$ for $u \in [U]$, $\mathbf{a} \in \mathcal{A}$.
\item[(ii)] $K=50$ cells and $U=3$ searchers. Cell means $\lambda_k$ are sampled from Uniform distributions on the intervals $[k,k+10]$ for $k=1,...,10$, $[20-k,30-k]$ for $k=11,...,20$, $[k-20,k-10]$ for $k=21,...,30$, $[40-k,50-k]$ for $k=31,...,40$, and $[k-40,k-30]$ for $k=41,...,50$. Baseline detection probabilities $\omega_{ku}$ are sampled from $\text{Beta}(u+2,2)$ distributions for $u \in [U]$, $k \in [K]$. Scaling functions are $\phi_u(\mathbf{a})=(0.5+0.5\sum_{k=1}^K \mathbb{I}\{a_k=u\})^{-1}$ for $u \in [U]$, $\mathbf{a} \in \mathcal{A}$.
\item[(iii)] $K=25$ cells and $U=10$ searchers. Cell means $\lambda_k$ are sampled from a $\text{Uniform}(90,100)$ distribution for $k \in [K]$. Baseline detection probabilities $\omega_{ku}$ are sampled from a $\text{Beta}(30,5)$ distribution for $u \in [U]$, $k \in [K]$. Scaling functions are $\phi_u(\mathbf{a})=(\sum_{k=1}^K \mathbb{I}\{a_k=u\})^{-1}$ for $u \in [U]$, $\mathbf{a} \in \mathcal{A}$.
\item[(iv)] $K=25$ cells and $U=5$ searchers. Cell means $\lambda_k$ are sampled from a $\text{Uniform}(0.4,1)$ distribution for $k \in [K]$. Baseline detection probabilities $\omega_{ku}$ are sampled from a $\text{Beta}(1,1)$ distribution for $u \in [U]$, $k \in [K]$. Scaling functions are $\phi_u(\mathbf{a})=(0.5+0.5\sum_{k=1}^K \mathbb{I}\{a_k=u\})^{-1}$ for $u \in [U]$, $\mathbf{a} \in \mathcal{A}$.
\end{enumerate}

We test a variety of parametrisations of {FP-CUCB} (in terms of $\lambda_{max}$) and TS (in terms of the prior mean and variance - from which particular $\alpha$ and $\beta$ values can be uniquely found) in each test. In each case we use $\lambda_{max}$ values which are both larger and smaller than the true maximal rate. Similarly we investigate TS with prior mean larger and smaller than the true maximal rate and with several different levels of variance. It is not always fully realistic to assume  knowledge of $\lambda_{max}$ will be perfect and therefore it is of interest to investigate the effects of varying it. Also, the choice of prior parameters in TS is a potentially subjective one and it is important to understand its impact.

We measure the performance of our algorithms by calculating the expected regret incurred by their actions, rescaled by the expected reward of a single optimal action. For an algorithm $A$ and particular history $\mathbf{H}_n$ we write \begin{displaymath}
ScaleReg_{\bs\lambda,\bs\gamma}^A(\mathbf{H}_n) = \frac{\sum_{t=1}^n \Delta_{\mathbf{a}_t}}{\text{opt}_{\bs\lambda,\bs\gamma}}.
\end{displaymath}
We calculate this value for all algorithms, all 250 datasets and rounds $1 \leq n \leq  2000$. We choose to rescale our regret to make a fairer comparison across the 50 different problem instances in each test (i)-(iv) which will all have different optimal expected rewards.

\begin{figure} 
	\includegraphics[width=\textwidth]{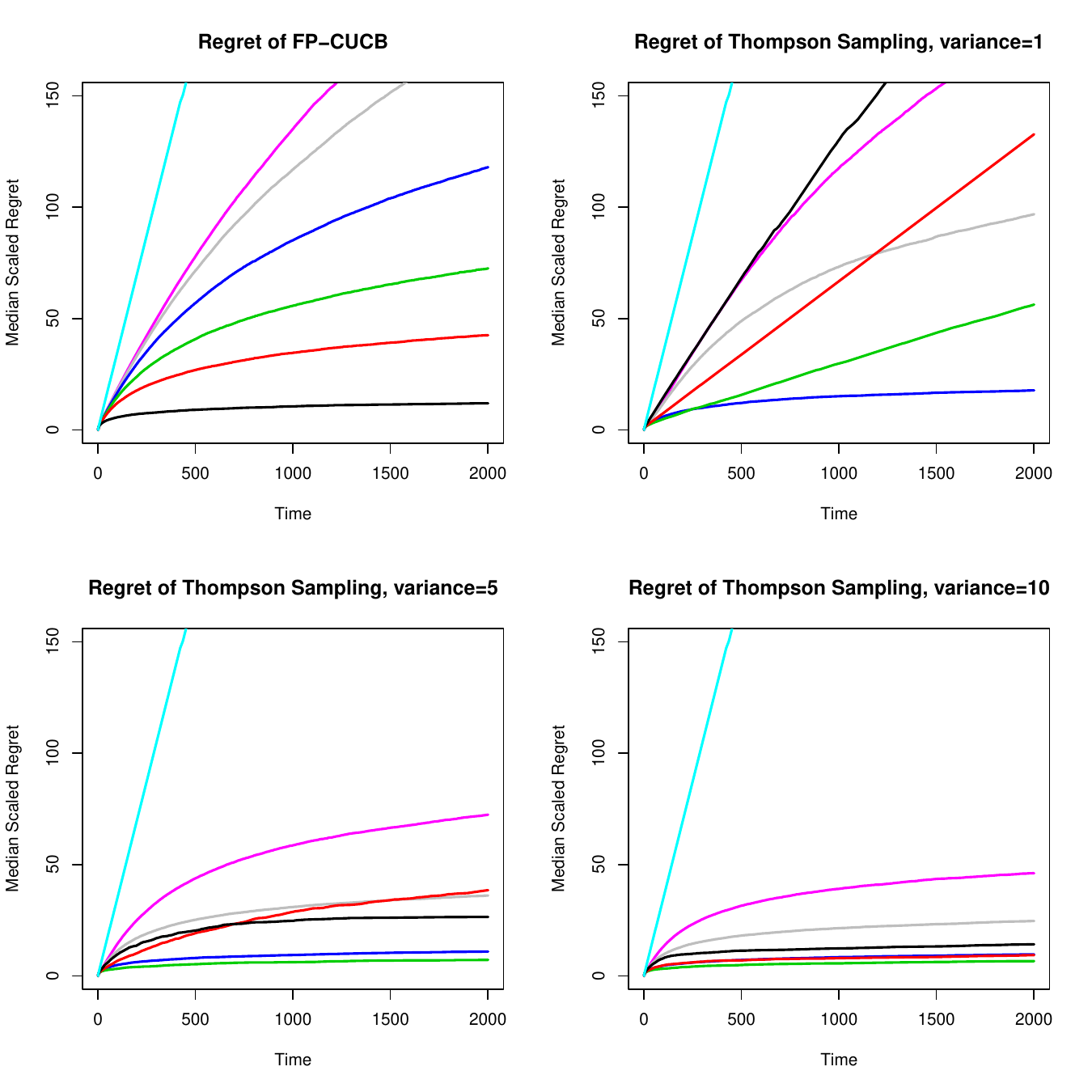}
	\caption{Cumulative Regret histories for Test (i). Upper left: FP-CUCB, upper right: TS with a prior variance of 1, lower left: TS with a prior variance of 5, lower right: TS with prior variance of 10. In each case the plotted lines are the median values of scaled regret calculated at each time point from 1 to 2000. Black lines represent $\lambda_{max}=1$ or a prior mean of 1, red represents the same parameters taking the value 5, green 10, blue 20, grey 40, and pink 60. In all sub-figures the teal line represents regret of the greedy algorithm.}
	\label{fig::test1}
\end{figure}

In Figure \ref{fig::test1} we illustrate how regret evolves over time by plotting the median scaled regret across the 250 runs of each algorithm in all rounds of test (i). The rate of growth shown in these plots is typical of the results in the other three tests. An immediate observation is that the greedy algorithm does very poorly on average and its full median regret over the 2000 rounds cannot be included in the graphs without obscuring differences between the other algorithms. We see also that the performance of both FP-CUCB and TS is strongly linked to the chosen parameters. For the FP-CUCB algorithm it seems in Figure 1 that the larger the parameter $\lambda_{max}$ is the larger the cumulative regret becomes. For TS, larger prior variances seem to induce lower regret, the relationship with the prior mean is more complex. Accurate specification of the prior mean seems to ensure good performance, but underestimation and overestimation of the mean can lead to poor performance (particularly when the variance is small).

\begin{figure}
	\includegraphics[width=0.5\linewidth]{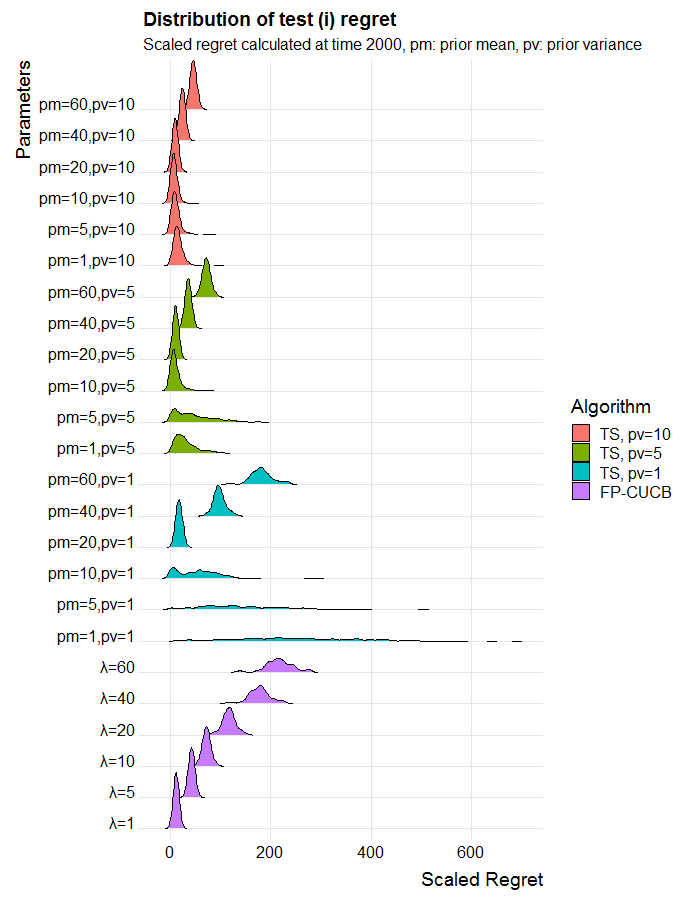}
	\includegraphics[width=0.5\linewidth]{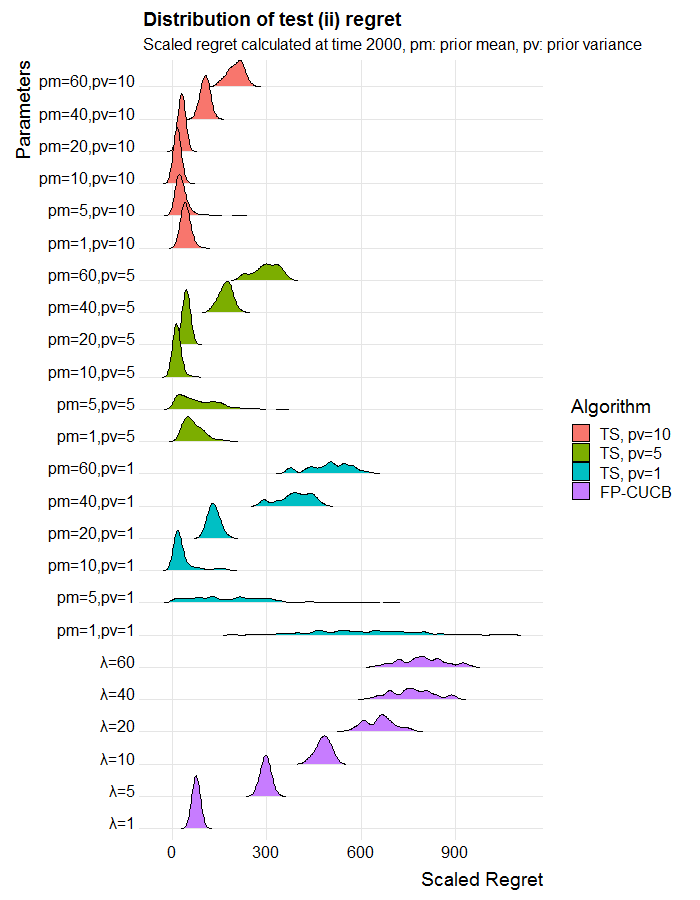}
	\caption{Scaled regret distributions in tests (i) and (ii). In both tests we have a true largest rate of 20.} \label{fig::ridge_i}
\end{figure}

\begin{figure}
	\includegraphics[width=0.5\linewidth]{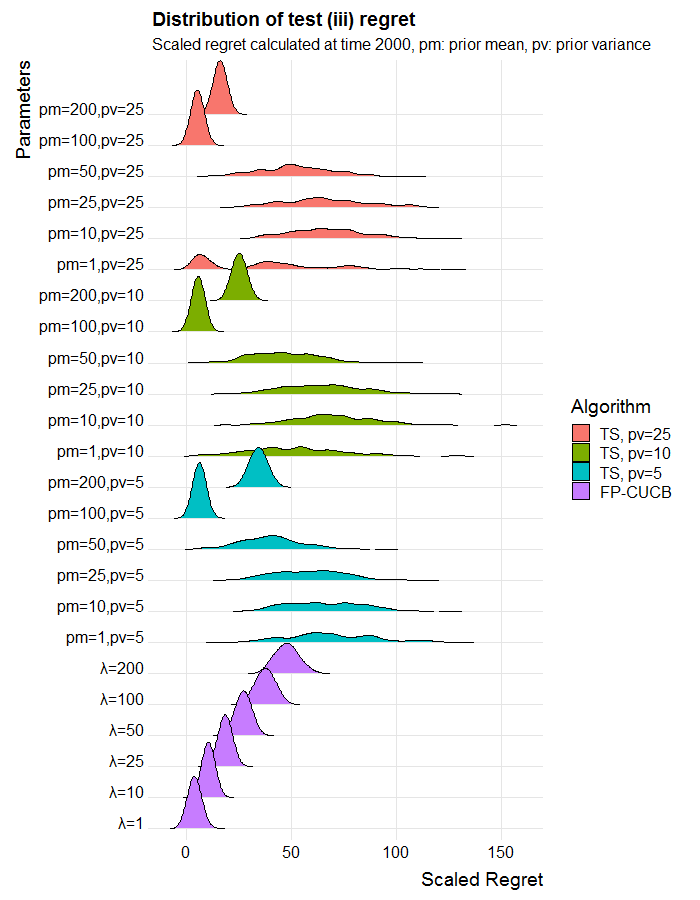}
	\includegraphics[width=0.5\linewidth]{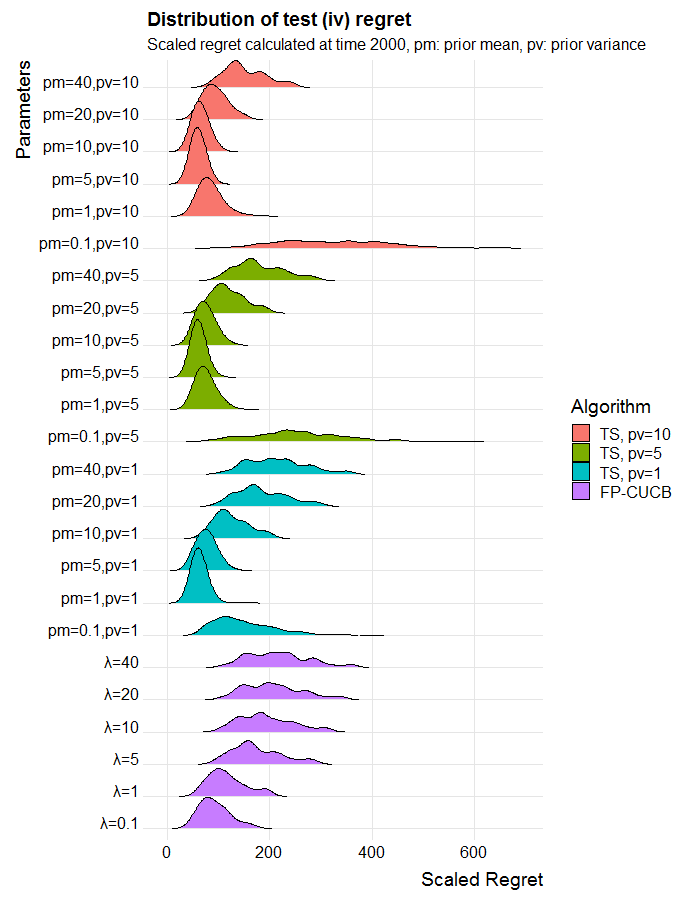}
	\caption{Scaled regret distributions in tests (iii) and (iv). In test (iii) the true largest rate is 100, and in test (iv) the true largest rate is 1.} \label{fig::ridge_ii}
\end{figure}

We analyse these behaviours further in Figures \ref{fig::ridge_i} and \ref{fig::ridge_ii}. Here we calculate a scaled regret at time $n=2000$ for all 250 runs of each algorithm and plot the empirical distribution of these values for each parameterisation of each algorithm. The results for tests (i) and (ii) are given in Figure \ref{fig::ridge_i} and for tests (iii) and (iv) in Figure \ref{fig::ridge_ii}. We omit the greedy algorithm's performance from these figures as the values are so large. In Appendix \ref{app::tables} we provide median values and lower and upper quantiles of the scaled regret for each algorithm. We see from these values that the greedy algorithm performs substantially worse than the FP-CUCB and TS algorithms which better address the exploration-exploitation dilemma. 

Examining Figures 2 and 3 we see that the FP-CUCB algorithm enjoys greater robustness to parameter choice than the TS approach. In particular in the results of test (iii) we see that many parametrisations of TS give rise to a long tailed distribution of round 2000 regret - meaning the performance of TS is highly variable and often poor. This variability of performance does seem to coincide with underestimation of the mean, however FP-CUCB manages to maintain strong performance even when the $\lambda_{max}$ parameter is far from the true maximal rate. When the prior variance is sufficiently large and the prior mean is close to the true $\lambda_{max}$ TS seems to do the best job of balancing exploration and exploitation and incurs the smallest regret.

\section{Discussion} \label{sec::discussion}
In this paper we have considered the problem of adaptively assigning multiple searchers to cells along a line (in space or time) in order to detect the maximum number of events occurring along the line. The problem is real, and has important applications in ecology, security, defence and other areas. We have modelled the problem, and proposed and analysed solution methods. The challenge at the heart of this problem is to correctly balance exploration and exploitation, in the face of initial ignorance as to the arrival process of events.

We formulated our sequential decision problem as a combinatorial multi-armed bandit with Poisson rewards and a novel filtered feedback mechanism. To design quality policies for this problem we first derived an efficient solution method to the full information problem. This IP forms the backbone of all policies for the sequential problem, as it allows us to quickly identify an optimal solution given some estimate of the arrival process' rate parameters.

We considered the sequential problem in two informational scenarios - firstly where the probability of detecting events is known, and secondly where these probabilities are unknown but one knows how they scale as the number of cells searched increases. For both of these cases we proposed an upper confidence bound approach. We derived lower bounds on the regret of all uniformly good algorithms under this our new feedback mechanism and upper bounds on the regret of our proposed approach.

In addition to the advantage of theoretical guarantees, the FP-CUCB algorithm is somewhat more reliable than TS. It is clear from the results of Section \ref{sec::numerics} that TS outperforms FP-CUCB for certain parametrisations (commonly larger choices of variance and mean close to the true arrival rates). However, we see that TS is particularly vulnerable to poor performance when the mean of the prior underestimates the true rate parameters. Even though our theoretical results for {FP-CUCB} depend on $\lambda_{max} \geq \lambda_k, k \in [K]$ we see that it is robust to underestimating this parameter. The reason FP-CUCB still performs well even when a key assumption does not hold is likely due to the fact that de la Pe\~{n}a's inequality does not give the tightest possible bound on Poisson tail probabilities (and therefore the rate of concentration of the mean). However, in order to construct the algorithm we required a symmetric tail bound for which an inflation term giving the type of concentration in Lemma 1 could be identified. Other bounds may be tighter but lack these properties. 

The variability of TS most likely arises due to the potential for the Gamma conjugate prior to be dominated by a small number of observations and create a scenario where TS behaves similarly to a greedy policy - sometimes fixing on good actions, but sometimes on poor ones. This phenomenon of variability of regret is understudied in multi-armed bandits, not least because it is much more challenging to analyse theoretically. However, in practical scenarios (where of course the learning and regret minimisation process will only occur once) this is a risk of TS. We note that both algorithms comfortably outperform the greedy algorithm in almost all examples, which speaks to the benefit of making some attempt to balance exploration and exploitation. 

An alternative treatment of bandit decision making is the \emph{non-stochastic} or \emph{adversarial} bandit \citep{AuerEtAl1995}. Under such a model, the assumptions that rewards are drawn i.i.d. from a fixed distribution are dropped, and may instead be any arbitrary sequence. Adversarial bandits necessitate a randomised strategy to guarantee good performance across any chosen reward sequence. Such methods have been developed in the MAB and CMAB settings \citep{AuerEtAl1995,CesaBianchiLugosi2012}. As further work the problem could be studied under a non-stochastic, or even a fully game-theoretic framework, relaxing some of our assumptions. This would however require a markedly different set of algorithmic and analytical tools. Within application domains, variants of the problem exist all along the spectrum from purely stochastic to fully game-theoretic. Our work has considered the stochastic setting in detail and in doing so provided a solution to many real-world problems.

\textbf{Acknowledgements} We gratefully acknowledge the support of the EPSRC funded EP/L015692/1 STOR-i Centre for Doctoral Training.

\bibliographystyle{apalike} 
\bibliography{orpaperrefs}

\begin{thebibliography}{}

\bibitem[Adams et~al., 2009]{Adams2009}
Adams, R.~P., Murray, I., and MacKay, D.~J. (2009).
\newblock Tractable nonparametric bayesian inference in poisson processes with
  gaussian process intensities.
\newblock In {\em Proceedings of the 26th Annual International Conference on
  Machine Learning}, pages 9--16. ACM.

\bibitem[Agrawal and Goyal, 2012]{Agrawal2012}
Agrawal, S. and Goyal, N. (2012).
\newblock Analysis of thompson sampling for the multi-armed bandit problem.
\newblock In {\em Conference on Learning Theory}, pages 39--1.

\bibitem[Anantharam et~al., 1987]{Anantharam1987i}
Anantharam, V., Varaiya, P., and Walrand, J. (1987).
\newblock Asymptotically efficient allocation rules for the multiarmed bandit
  problem with multiple plays-part i: Iid rewards.
\newblock {\em IEEE Transactions on Automatic Control}, 32(11):968--976.

\bibitem[Auer et~al., 2002]{Auer2002}
Auer, P., Cesa-Bianchi, N., and Fischer, P. (2002).
\newblock {Finite-Time Analysis of the Multiarmed Bandit Problem}.
\newblock {\em Machine Learning}, 47(2-3):235--256.

\bibitem[Auer et~al., 1995]{AuerEtAl1995}
Auer, P., Cesa-Bianchi, N., Freund, Y., and Schapire, R.~E. (1995).
\newblock Gambling in a rigged casino: The adversarial multi-armed bandit
  problem.
\newblock In {\em Proceedings of IEEE 36th Annual Foundations of Computer
  Science}, pages 322--331. IEEE.

\bibitem[Benes, 1957]{Benes1957}
Benes, V. (1957).
\newblock On queues with poisson arrivals.
\newblock {\em The Annals of Mathematical Statistics}, pages 670--677.

\bibitem[Bubeck and Cesa-Bianchi, 2012]{Bubeck2012}
Bubeck, S. and Cesa-Bianchi, N. (2012).
\newblock {Regret Analysis of Stochastic and Nonstochastic Multi-Armed Bandit
  Problems}.
\newblock {\em {In Foundations and Trends in Machine Learning}}, 5(1):1--122.

\bibitem[Bubeck et~al., 2013]{Bubeck2013}
Bubeck, S., Cesa-Bianchi, N., and Lugosi, G. (2013).
\newblock {Bandits with heavy tail}.
\newblock {\em {IEEE Transactions on Information Theory}}, 59(11):7711--7717.

\bibitem[Burnetas and Katehakis, 1996]{Burnetas1996optimal}
Burnetas, A.~N. and Katehakis, M.~N. (1996).
\newblock Optimal adaptive policies for sequential allocation problems.
\newblock {\em Advances in Applied Mathematics}, 17(2):122--142.

\bibitem[Capp{\'e} et~al., 2013]{Cappe2013}
Capp{\'e}, O., Garivier, A., Maillard, O.-A., Munos, R., and Stoltz, G. (2013).
\newblock {Kullback--Leibler upper confidence bounds for optimal sequential
  allocation}.
\newblock {\em {The Annals of Statistics}}, 41(3):1516--1541.

\bibitem[Carlsson et~al., 2016]{Carlsson2013}
Carlsson, J.~G., Carlsson, E., and Devulapalli, R. (2016).
\newblock {Shadow Prices in Territory Division}.
\newblock {\em Networks and Spatial Economics}, 16(3):893--931.

\bibitem[Cesa-Bianchi and Lugosi, 2012]{CesaBianchiLugosi2012}
Cesa-Bianchi, N. and Lugosi, G. (2012).
\newblock Combinatorial bandits.
\newblock {\em Journal of Computer and System Sciences}, 78(5):1404--1422.

\bibitem[Chen et~al., 2016a]{Chen2016a}
Chen, W., Hu, W., Li, F., Li, J., Liu, Y., and Lu, P. (2016a).
\newblock Combinatorial multi-armed bandit with general reward functions.
\newblock In {\em Advances in Neural Information Processing Systems}, pages
  1651--1659.

\bibitem[Chen et~al., 2013]{Chen2013}
Chen, W., Wang, Y., and Yuan, Y. (2013).
\newblock {Combinatorial multi-armed bandit: General framework and
  applications}.
\newblock In {\em {Proceedings of the 30th International Conference on Machine
  Learning}}, pages 151--159.

\bibitem[Chen et~al., 2016b]{Chen2016b}
Chen, W., Wang, Y., Yuan, Y., and Wang, Q. (2016b).
\newblock Combinatorial multi-armed bandit and its extension to
  probabilistically triggered arms.
\newblock {\em Journal of Machine Learning Research}, 17(50):1--33.

\bibitem[Combes et~al., 2015]{Combes2015}
Combes, R., Shahi, M. S. T.~M., Proutiere, A., and Lelarge, M. (2015).
\newblock {Combinatorial bandits revisited}.
\newblock In {\em {Advances in Neural Information Processing Systems}}, pages
  2116--2124.

\bibitem[Cowan et~al., 2015]{Cowan2015Normal}
Cowan, W., Honda, J., and Katehakis, M.~N. (2015).
\newblock Normal bandits of unknown means and variances: Asymptotic optimality,
  finite horizon regret bounds, and a solution to an open problem.
\newblock {\em arXiv preprint arXiv:1504.05823}.

\bibitem[de~la Pe\~{n}a, 1999]{DeLaPena1999}
de~la Pe\~{n}a, V. (1999).
\newblock A general class of exponential inequalities for martingales and
  ratios.
\newblock {\em The Annals of Probability}, 27(1):537--564.

\bibitem[Diggle, 1985]{Diggle1985}
Diggle, P. (1985).
\newblock A kernel method for smoothing point process data.
\newblock {\em Applied statistics}, pages 138--147.

\bibitem[Gai et~al., 2012]{Gai2012}
Gai, Y., Krishnamachari, B., and Jain, R. (2012).
\newblock {Combinatorial network optimization with unknown variables:
  Multi-armed bandits with linear rewards and individual observations}.
\newblock {\em {IEEE/ACM Transactions on Networking (TON)}}, 20(5):1466--1478.

\bibitem[Garey and Johnson, 1979]{Garey1979}
Garey, M.~R. and Johnson, D.~S. (1979).
\newblock Computers and intractability: a guide to np-completeness.

\bibitem[Garivier and Capp{\'e}, 2011]{Garivier2011}
Garivier, A. and Capp{\'e}, O. (2011).
\newblock {The KL-UCB Algorithm for Bounded Stochastic Bandits and Beyond}.
\newblock In {\em COLT}, pages 359--376.

\bibitem[Gittins et~al., 2011]{Gittins2011}
Gittins, J., Glazebrook, K., and Weber, R. (2011).
\newblock {\em Multi-Armed Bandit Allocation Indices}.
\newblock John Wiley \& Sons.

\bibitem[Graves and Lai, 1997]{Graves1997}
Graves, T.~L. and Lai, T.~L. (1997).
\newblock Asymptotically efficient adaptive choice of control laws incontrolled
  markov chains.
\newblock {\em SIAM Journal on Control and Optimization}, 35(3):715--743.

\bibitem[Gugushvili et~al., 2018]{Gugushvili2018}
Gugushvili, S., van~der Meulen, F., Schauer, M., and Spreij, P. (2018).
\newblock Fast and scalable non-parametric bayesian inference for poisson point
  processes.
\newblock {\em arXiv preprint arXiv:1804.03616}.

\bibitem[Heikkinen and Arjas, 1999]{Heikkinen1999}
Heikkinen, J. and Arjas, E. (1999).
\newblock Modeling a {Poisson} forest in variable elevations: a nonparametric
  bayesian approach.
\newblock {\em Biometrics}, 55:738--745.

\bibitem[Helmers et~al., 2005]{Helmers2005}
Helmers, R., Mangku, I.~W., and Zitikis, R. (2005).
\newblock Statistical properties of a kernel-type estimator of the intensity
  function of a cyclic poisson process.
\newblock {\em Journal of Multivariate Analysis}, 92(1):1--23.

\bibitem[John and Hensman, 2018]{John2018}
John, S. and Hensman, J. (2018).
\newblock Large-scale cox process inference using variational fourier features.
\newblock {\em arXiv preprint arXiv:1804.01016}.

\bibitem[Kaufmann, 2016]{Kaufmann2016}
Kaufmann, E. (2016).
\newblock {On Bayesian index policies for sequential resource allocation}.
\newblock {\em arXiv preprint arXiv:1601.01190}.

\bibitem[Kaufmann et~al., 2012]{Kaufmann2012thompson}
Kaufmann, E., Korda, N., and Munos, R. (2012).
\newblock Thompson sampling: An asymptotically optimal finite-time analysis.
\newblock In {\em International Conference on Algorithmic Learning Theory},
  pages 199--213. Springer.

\bibitem[Kirichenko and Van~Zanten, 2015]{Kirichenko2015}
Kirichenko, A. and Van~Zanten, H. (2015).
\newblock Optimality of poisson processes intensity learning with gaussian
  processes.
\newblock {\em The Journal of Machine Learning Research}, 16(1):2909--2919.

\bibitem[Koopman, 1946]{koopman1946search}
Koopman, B. (1946).
\newblock Search and screening, operations evaluation group report 56.
\newblock {\em Center for Naval Analysis, Alexandria, Virginia}.

\bibitem[Kveton et~al., 2015a]{Kveton2015cascading}
Kveton, B., Szepesvari, C., Wen, Z., and Ashkan, A. (2015a).
\newblock Cascading bandits: Learning to rank in the cascade model.
\newblock In {\em International Conference on Machine Learning}, pages
  767--776.

\bibitem[Kveton et~al., 2015b]{Kveton2015tight}
Kveton, B., Wen, Z., Ashkan, A., and Szepesvari, C. (2015b).
\newblock Tight regret bounds for stochastic combinatorial semi-bandits.
\newblock In {\em Artificial Intelligence and Statistics}, pages 535--543.

\bibitem[Lai and Robbins, 1985]{Lai1985asymptotically}
Lai, T.~L. and Robbins, H. (1985).
\newblock Asymptotically efficient adaptive allocation rules.
\newblock {\em Advances in Applied Mathematics}, 6(1):4--22.

\bibitem[Lattimore, 2017]{Lattimore2017}
Lattimore, T. (2017).
\newblock A scale free algorithm for stochastic bandits with bounded kurtosis.
\newblock In {\em Advances in Neural Information Processing Systems}, pages
  1583--1592.

\bibitem[Luedtke et~al., 2016]{Luedtke2016}
Luedtke, A., Kaufmann, E., and Chambaz, A. (2016).
\newblock Asymptotically optimal algorithms for multiple play bandits with
  partial feedback.
\newblock {\em arXiv preprint arXiv:1606.09388}.

\bibitem[Riordan, 1937]{Riordan1937}
Riordan, J. (1937).
\newblock Moment recurrence relations for binomial, poisson and hypergeometric
  frequency distributions.
\newblock {\em The Annals of Mathematical Statistics}, 8(2):103--111.

\bibitem[Russo and Van~Roy, 2016]{Russo2016}
Russo, D. and Van~Roy, B. (2016).
\newblock An information-theoretic analysis of thompson sampling.
\newblock {\em The Journal of Machine Learning Research}, 17(1):2442--2471.

\bibitem[Serra et~al., 2014]{Serra2014}
Serra, L., Saez, M., Juan, P., Varga, D., and Mateu, J. (2014).
\newblock A spatio-temporal poisson hurdle point process to model wildfires.
\newblock {\em Stochastic environmental research and risk assessment},
  28(7):1671--1684.

\bibitem[Stone, 1976]{Stone1976}
Stone, L.~D. (1976).
\newblock {\em Theory of Optimal Search}, volume 118.
\newblock Elsevier.

\bibitem[Szechtman et~al., 2008]{Szechtman2008}
Szechtman, R., Kress, M., Lin, K., and Cfir, D. (2008).
\newblock {Models of Sensor Operations for Border Surveillance}.
\newblock {\em Naval Research Logistics (NRL)}, 55(1):27--41.

\bibitem[Thompson, 1933]{Thompson1933}
Thompson, W.~R. (1933).
\newblock On the likelihood that one unknown probability exceeds another in
  view of the evidence of two samples.
\newblock {\em Biometrika}, 25(3/4):285--294.

\bibitem[Wang and Chen, 2018]{Wang2018}
Wang, S. and Chen, W. (2018).
\newblock Thompson sampling for combinatorial semi-bandits.
\newblock arXiv:1803.04623.

\bibitem[Washburn, 2002]{Washburn2002}
Washburn, A.~R. (2002).
\newblock {\em Search and Detection}.
\newblock INFORMS.

\bibitem[Weinberg et~al., 2007]{Weinberg2007}
Weinberg, J., Brown, L.~D., and Stroud, J.~R. (2007).
\newblock Bayesian forecasting of an inhomogeneous poisson process with
  applications to call center data.
\newblock {\em Journal of the American Statistical Association},
  102(480):1185--1198.

\end{thebibliography}

\appendix

\section{Proof of NP-hardness of the IP (\ref{eq::IP})} \label{app:NPhard}

\begin{theorem} \label{thm::NPhard}
Integer Linear Programs of the following type are $\mathcal{NP}$-hard in the strong sense: \begin{align*}
\max_{a_{iju}, 1 \leq i \leq j \leq K, u \in [U]} \sum_{i=1}^K \sum_{j=i}^K \sum_{u=1}^U q_{iju}a_{iju} & \\
\text{such that } \sum_{i=1}^K \sum_{j=i}^K a_{iju} \leq &1, \enspace u \in [U] \\
\sum_{i=1}^K \sum_{j=i}^K \sum_{u=1}^U a_{iju} \leq &1, \enspace k \in [K] \\
a_{iju} \in &\{0,1\}, \enspace 1 \leq i \leq j \leq K, u \in [U].
\end{align*}
\end{theorem}

\noindent
\emph{Proof of Theorem \ref{thm::NPhard}:}

The following problem is known to be $\mathcal{NP}$-complete in the strong sense \citep{Garey1979}: 

\noindent \texttt{3-PARTITION}: Given positive integers $w_1,...,w_{3n}$ and a positive integer ``target" $t$, does there exist a partition of $\{1,...,3n\}$ into subsets $S_1,...,S_n$ such that $|S_i|=3$ and $\sum_{j \in S_i} w_j=t$ for $i=1,...,n$?

We reduce this to an IP of the given type as follows. First, we assume without loss of generality that $\sum_{j=1}^{3n} w_j=nt$, since otherwise the answer to \texttt{3-PARTITION} is trivially ``no". Let $K=nt$ and $U=3n$. For $k=1,...,3n$, set $q_{iju}=w_u$ if $j-i=w_u$ and the half-open interval $[i,j)$ does not include a multiple of $t$. Set all other $q_{iju}$ to zero. Then the answer to \texttt{3-PARTITION} is ``yes" if and only if there is a solution to the IP with profit equal to $nt$. $\square$

\section{Lemma \ref{lem::FPCUCBconc} Proof: Concentration of filtered Poisson estimator} \label{proof::concresult}
By definition $Z_j = \sum_{i=1}^j (Y_i - \mathbb{E}(Y_i))$, the sum of the accumulated noise to round $j$ is a martingale. Therefore, $W_j = Z_j - Z_{j-1}= \sum_{i=1}^j (Y_i - \mathbb{E}(Y_i)) - \sum_{i=1}^{j-1} (Y_i - \mathbb{E}(Y_i))= Y_j - \mathbb{E}(Y_j)$ is a martingale difference sequence. We will utilise the following concentration result for martingale difference sequences due to \cite{DeLaPena1999}: \begin{theorem}[de la Pe\~{n}a's inequality]
	Let $\{d_i,\mathcal{F}_i\}$ be a martingale difference sequence with $\mathbb{E}(d_j|\mathcal{F}_{j=1})=0$, $\mathbb{E}(d_j^2|\mathcal{F}_{j-1})=\sigma_j^2$, $V_n^2 = \sum_{j=1}^n \sigma_j^2$. Furthermore assume that $\mathbb{E}(|d_j|^k|\mathcal{F}_{j-1}) \leq \frac{k!}{2} \sigma_j^2 c^{k-2}$ for $k \geq 2, 0< c < \infty$. Then, for all $x,y>0$ \begin{equation}
	\mathbb{P}\bigg(\sum_{i=1}^n d_i \geq x, V_n^2 \leq y \text{ for some } n\bigg) \leq \exp \bigg(\frac{-x^2}{2(y+cx)}\bigg).
	\end{equation}
\end{theorem}
Plainly, $\mathbb{E}(W_j|\cdot)=0$ and $\mathbb{E}(W_j^2|\cdot)=\gamma_j\lambda$. The proof of the condition on higher order moments is more involved. Firstly we define $\mu_k$ to be the $k^{th}$ central moment of a Poisson distribution with parameter $\lambda$. \cite{Riordan1937} gives us the following second order recurrence relationship for the central moments $\mu$ of the Poisson distribution  \begin{equation*}
\mu_{k} = \lambda\bigg( \frac{d\mu_{k-1}}{d\lambda} + (k-1)\mu_{k-2}\bigg), \quad k=2,3,...
\end{equation*} 
We first demonstrate a bound on the order (with respect to $\lambda$) of $\mu_k$.

\begin{lemma} \label{lem::momentlemma}
	For $k \geq 2$, $\mu_k=o(\lambda^{k/2})$.
\end{lemma}

\noindent
\emph{Proof of Lemma \ref{lem::momentlemma}} We can prove Lemma \ref{lem::momentlemma} via an induction argument. Note that $\mu_1=0$ and $\mu_2=\lambda$. Assume for some $r>3$ that $\mu_r=o(\lambda^{r/2})$ and $\mu_{r-1}=o(\lambda^{(r-1)/2})$. Then consider $\mu_{r+1}= \lambda \frac{d\mu_r}{d\lambda}+ r\lambda\mu_{r-1}$. For the first term we have $\frac{d\mu_r}{d\lambda}=o(\lambda^{r/2 -1})$ and thus $\lambda\frac{d\mu_r}{d\lambda}=o(\lambda^{r/2})$. The second term is plainly of order $o(\lambda^{(r+1)/2})$ and thus $\mu_{r+1}=o(\lambda^{(r+1)/2})$, completing the induction argument and proving Lemma \ref{lem::momentlemma} $\square$.

Now introduce $\nu_k=\frac{k!}{2}\lambda\max(1,\sqrt{\lambda})^{k-2}$ for $k\geq 2$. The following lemma will be sufficient to demonstrate that the condition on higher order moments holds.
\begin{lemma} \label{lem::bernsteinlemma}
	For $k \geq 2$, $\nu_k \geq \mu_k$.
\end{lemma}

\noindent
\emph{Proof of Lemma \ref{lem::bernsteinlemma}} Firstly we write $\nu_k$ as a recurrence relationship \begin{equation*}
\nu_k = k\max(1,\sqrt{\lambda})\nu_{k-1} = k(k-1)\max(1,\sqrt{\lambda})^2\nu_{k-2}, \quad k=2,3,... 
\end{equation*} We also prove this Lemma via an induction argument, which proceeds as follows. For $\mu_k$ we have the following initial values $\mu_2=\lambda$, $\mu_3=\lambda$, $\mu_4=4\lambda^2$ and for $\nu_k$ we have $\nu_2=\lambda$, $\nu_3 = 3\lambda\max(1,\sqrt{\lambda})$, $\nu_4=12\lambda\max(1,\sqrt{\lambda})^2$. Clearly, these initial values satisfy $\nu_k \geq \mu_k$. Now assume that for some $p > 5$, we have $\nu_p \geq \mu_p$ and $\nu_{p-1} \geq \mu_{p-1}$. Then consider $\mu_{p+1}$ as follows: \begin{align*}
\mu_{p+1} 	&= \lambda \frac{d\mu_p}{d\lambda} + p\lambda\mu_{p-1} \\
&\leq \lambda \frac{d\mu_p}{d\lambda} + p\lambda \nu_{p-1} \\
&\leq \frac{p}{2} \mu_p + p\lambda\nu_{p-1} \\
&\leq \frac{p}{2} \nu_p + p\lambda\nu_{p-1} \\
&= \frac{p^2}{2} \max(1,\sqrt{\lambda}) \nu_{p-1} + p\lambda\nu_{p-1} \\
&\leq \max(1,\sqrt{\lambda})^2(\frac{1}{2}p^2+p)\nu_{p-1} \\
&\leq \max(1,\sqrt{\lambda})^2(p+1)p\nu_{p-1} = \nu_{p+1},
\end{align*} completing the proof by induction. The first and third inequalities are due to the assumed relationships for $p$ and $p-1$, the second inequality is a consequence of Lemma \ref{lem::momentlemma} and the differentiation of a polynomial. $\square$.

The martingale difference sequence $W_j$ therefore satisfies the conditions of de la Pe\~{n}a's inequality with $c=\max(1,\sqrt{\lambda})$ and we have \begin{equation*}
\mathbb{P}\bigg(\sum_{j=1}^s W_j \geq x, \sum_{j=1}^s \mathbb{E}(W_j^2|\cdot) \leq y\bigg) \leq \exp \bigg(\frac{-x^2}{2y + 2\max(1,\sqrt{\lambda})x}\bigg).
\end{equation*} We have that $\sum_{j=1}^s \mathbb{E}(W_j^2|\cdot) \leq \sum_{j=1}^s \gamma_j\lambda$ with probability 1, so we may use the simplified result \begin{equation*}
\mathbb{P}\bigg(\sum_{j=1}^s W_j \geq x\bigg) \leq \exp \bigg(\frac{-x^2}{2\lambda\sum_{j=1}^s \gamma_j + 2\max(1,\sqrt{\lambda})x}\bigg).
\end{equation*} Then if $x= 6\max(1,\sqrt{\lambda_{max}})\log(t) + \sqrt{6\lambda_{max}\sum_{j=1}^s \gamma_j \log(t) }$ , we have,
\begin{align*}
&P\bigg( \sum_{i=1}^s (Y_i - \mathbb{E}(Y_i)) > x \bigg) \\
&\leq \exp \Bigg(-\frac{36\max(1,\lambda_{max})\log^2(t)+ 12\max(1,\sqrt{\lambda_{max}})\log(t)\sqrt{6\lambda_{max}\sum_{j=1}^s \gamma_j \log(t) }+ 6\lambda_{max}\sum_{j=1}^s \gamma_j \log(t)}{2\lambda\sum_{j=1}^s \gamma_j +12\max(1,\sqrt{\lambda\cdot\lambda_{max}})\log(t)+ 2\max(1,\sqrt{\lambda})\sqrt{6\lambda_{max}\sum_{j=1}^s \gamma_j \log(t)}}\Bigg) \\
&= \exp \bigg( -\log(t) \frac{36\max(1,\lambda_{max})\log(t)+12\max(1,\sqrt{\lambda_{max}})\sqrt{6\lambda_{max}\sum_{j=1}^s \gamma_j\log(t)} + 6\lambda_{max}\sum_{j=1}^s \gamma_j}{12\max(1,\sqrt{\lambda\cdot\lambda_{max}})\log(t) + 2\max(1,\sqrt\lambda)\sqrt{6\lambda_{max}\sum_{j=1}^s \gamma_j\log(t)} + 2\lambda\sum_{j=1}^s \gamma_j} \bigg) \\
&= \exp \bigg( -3\log(t) \frac{12\max(1,\lambda_{max})\log(t)+4\max(1,\sqrt{\lambda_{max}})\sqrt{6\lambda_{max}\sum_{j=1}^s \gamma_j\log(t)} + 2\lambda_{max}\sum_{j=1}^s \gamma_j}{12\max(1,\sqrt{\lambda\cdot\lambda_{max}})\log(t) + 2\max(1,\sqrt\lambda)\sqrt{6\lambda_{max}\sum_{j=1}^s \gamma_j\log(t)} + 2\lambda\sum_{j=1}^s \gamma_j} \bigg) \leq t^{-3}.
\end{align*}
It follows that \begin{align*}
&P\bigg( \sum_{i=1}^s (Y_i - \mathbb{E}(Y_i)) > 6\max(1,\sqrt{\lambda_{max}})\log(t) + \sqrt{6\lambda_{max}\sum_{j=1}^s \gamma_j \log(t) } \bigg) \\
&= P\bigg( \sum_{i=1}^s Y_i - \lambda\sum_{i=1}^s \gamma_i  > 6\max(1,\sqrt{\lambda_{max}})\log(t) + \sqrt{6\lambda_{max}\sum_{j=1}^s \gamma_j \log(t) } \bigg) \\
&= P\bigg( \frac{\sum_{i=1}^s Y_i}{\sum_{i=1}^s \gamma_i} - \lambda > \frac{6\max(1,\sqrt{\lambda_{max}})\log(t) + \sqrt{6\lambda_{max}\sum_{j=1}^s \gamma_j \log(t) }}{\sum_{i=1}^n \gamma_i} \bigg) \leq t^{-3}.
\end{align*} Finally, note that $\bar{Z}_j=-Z_j= \sum_{i=1}^j (\mathbb{E}(Y_i)-Y_i)$ is also a martingale whose difference series satisfies the conditions of de la Pe\~{n}a's inequality and thus we can achieve the same bound for deviations on the left, and introduce achieve the required result. $\square$

\section{Theorem \ref{thm::FPCUCB} Proof: Expected regret of FP-CUCB} \label{Theorem1Proof}
To complete the proof of Theorem \ref{thm::FPCUCB} provided in the main text, we separately prove Propositions \ref{prop::FPCUCBsuff} and \ref{prop::FPCUCBund}.

\noindent \emph{Proof of Proposition \ref{prop::FPCUCBsuff}:} 

Here we prove a bound on the expected number of plays of an arm \emph{after} it has reached its sufficient sampling level. Define the event \begin{displaymath}
\mathcal{N}_t = \Bigg\{\bigg|\frac{\sum_{j=1}^{t-1}Y_{k,j}}{D_{k,t-1}}-\lambda_k\bigg| < \frac{6\max(1,\sqrt{\lambda_{max}})\log(t)}{D_{k,t-1}} + \sqrt{\frac{6\lambda_{max}\log(t)}{D_{k,t-1}}} \enspace \forall k\in[K]\Bigg\}.
	\end{displaymath} 
Define random variables $\Lambda_{k,t}=\frac{6\max(1,\sqrt{\lambda_{max}})\log(t)}{D_{k,t-1}} + \sqrt{\frac{6\lambda_{max}\log(t)}{D_{k,t-1}}}$ for $k\in[K]$ and $\Lambda_t=\max_{k:g_{k,t}>0}(\Lambda_{k,t})$. Define $\Lambda^{k,l}_t=\frac{6\max(1,\sqrt{\lambda_{max}})\log(t)}{\gamma_{k,min}h_{k,n}(\Delta^{k,l})} + \sqrt{\frac{6\lambda_{max}\log(t)}{\gamma_{k,min}h_{k,n}(\Delta^{k,l})}}$ for $l \in [B_k]$, $k \in [K]$, $t \in [n]$, which are not random variables. By these definitions and the definition of UCB indices $\bar{\lambda}_{k,t}$ we have the following properties.
\begin{align*}
&\mathcal{N}_t \Rightarrow  \bar{\lambda}_{k,t}-\lambda_k >0 \enspace \forall k \in [K]  \\
&\mathcal{N}_t \Rightarrow  \bar{\lambda}_{k,t}-\lambda_k < 2\Lambda_t \enspace \forall k: g_{k,t}>0 \\
&\{\mathbf{g}_t = \mathbf{g}_{k,B}^l, N_{k,t}>N_{k,t-1},  N_{s,t-1}> h_{k,n}(\Delta^{k,l}) \enspace \forall s:g_{s,t}>0\} \Rightarrow \Lambda^{k,l}_t> \Lambda_t  \enspace \forall k \in [K], \forall l \in [B_k] 
\end{align*}
For any particular $k \in [K]$ and $l \in [B_k]$ if $\{\mathcal{N}_t,\mathbf{g}_t = \mathbf{g}_{k,B}^l, N_{k,t}>N_{k,t-1},  N_{s,t-1}> h_{k,n}(\Delta^{k,l}) \enspace \forall s: g_{s,t}>0\}$ holds at time $t$ the following is implied
\begin{equation}
\mathbf{g}_t^T\cdot \bs\lambda + 2K\Lambda^{k,l}_t > \mathbf{g}_t^T\cdot \bs\lambda + 2K\Lambda_t \geq \mathbf{g}_t^T \cdot \bar{\bs\lambda}_t \geq (\mathbf{g}_{\bs\lambda}^*)^T \cdot \bar{\bs\lambda}_t\geq (\mathbf{g}_{\bs\lambda}^*)^T \cdot {\bs\lambda} = \text{opt}_{\bs\lambda,\bs\gamma} \label{Contradiction}
\end{equation}
where $\mathbf{g}^*_{\bs\lambda}$ is an action that is optimal with respect to rate vector $\bs\lambda$. However, by definition $2K\Lambda^{k,l}_t\geq\Delta^{k,l}$ and therefore (\ref{Contradiction}) is a contradiction of the definition of $\Delta^{k,l}=\text{opt}_{\bs\lambda,\bs\gamma}-\mathbf{g}_{k,B}^l\cdot \bs\lambda$. Therefore \begin{align*}
&\mathds{P}(\mathcal{N}_t,\mathbf{g}_t = \mathbf{g}_{k,B}^l, N_{k,t}>N_{k,t-1}, \forall s: g_{s,t}>0, N_{s,t-1}> h_{k,n}(\Delta^{k,l}))=0 \enspace \forall k \in [K], \enspace \forall l \in [B_k] 
\end{align*}
and \begin{displaymath}
\sum_{k=1}^K \sum_{l=1}^{B_k}\mathds{P}(\mathbf{g}_t = \mathbf{g}_{k,B}^l, N_{k,t}>N_{k,t-1},  N_{s,t-1}> h_{k,n}(\Delta^{k,l}) \enspace \forall s:g_{s,t}>0)\leq \mathds{P}(\neg \mathcal{N}_t) \leq 2Kt^{-2}.
\end{displaymath}
The bound on $\mathds{P}(\neg \mathcal{N}_t)$ comes from applying Lemma 1 and is sufficient to prove Proposition \ref{prop::FPCUCBsuff} since \begin{align*}
\mathds{E}\Bigg(\sum_{k=1}^K \sum_{l=1}^{B_k} N_{k,n}^{l,suf}\Bigg)&=\mathds{E}\Bigg(\sum_{t=K+1}^n \sum_{k=1}^K \sum_{l=1}^{B_k} \mathds{I}\{\mathbf{g}_t = \mathbf{g}_{k,B}^l, N_{k,t} > N_{k,t-1},N_{k,t-1}> h_{k,n}(\Delta^{k,l})\}\Bigg) \\
&\leq \sum_{t=K+1}^n 2Kt^{-2} \leq \frac{\pi^2}{3}\cdot K. \quad \square
\end{align*}

\noindent
\emph{Proof of Proposition 2}

Now consider the number of plays made prior to reaching the sufficient sampling level. Firstly set $h_{k,n}(\Delta^{k,0})=0$ to simplify notation and consider the following steps. Then for any cell $k$ in $\{j \in [K]|\Delta_{min}^j >0 \}$ \begin{align*}
\sum_{l=1}^{B_k} N_{k,n}^{l,und}\cdot \Delta^{k,l} &= \sum_{t=K+1}^n \sum_{l=1}^{B_k} \mathds{I}\bigg\{\mathbf{g}_t = \mathbf{g}_{k,B}^l, N_{k,t}> N_{k,t-1}, N_{k,t-1}\leq h_{k,n}(\Delta^{k,l})\bigg\} \Delta^{k,l} \\
&= \sum_{t=K+1}^n \sum_{l=1}^{B_k} \sum_{j=1}^l \mathds{I}\bigg\{ \mathbf{g}_t = \mathbf{g}_{k,B}^l, N_{k,t}> N_{k,t-1}, N_{k,t-1}\in \Big(h_{k,n}(\Delta^{k,j-1}),h_{k,n}(\Delta^{k,j})\Big) \bigg\} \Delta^{k,l} \\
&\leq \sum_{t=K+1}^n \sum_{l=1}^{B_k} \sum_{j=1}^l \mathds{I}\bigg\{\mathbf{g}_t = \mathbf{g}_{k,B}^l, N_{k,t}> N_{k,t-1}, N_{k,t-1}\in \Big(h_{k,n}(\Delta^{k,j-1}),h_{k,n}(\Delta^{k,j})\Big) \bigg\} \Delta^{k,j} \\
\intertext{as $\Delta^{k,1} \geq \Delta^{k,2} \geq ... \geq \Delta^{k,B_k}$,}
&\leq \sum_{t=K+1}^n \sum_{l=1}^{B_k} \sum_{j=1}^{B_k} \mathds{I}\bigg\{\mathbf{g}_t = \mathbf{g}_{k,B}^l, N_{k,t}> N_{k,t-1}, N_{k,t-1}\in \Big(h_{k,n}(\Delta^{k,j-1}),h_{k,n}(\Delta^{k,j})\Big) \bigg\} \Delta^{k,j} \\
&= \sum_{t=K+1}^n \sum_{j=1}^{B_k} \mathds{I}\bigg\{ \mathbf{g}_t \in \mathcal{G}_{k,B}, N_{k,t}> N_{k,t-1}, N_{k,t-1}\in \Big(h_{k,n}(\Delta^{k,j-1}),h_{k,n}(\Delta^{k,j})\Big) \bigg\} \Delta^{k,j} \\
&= \sum_{j=1}^{B_k} \sum_{t=K+1}^n  \mathds{I}\bigg\{\mathbf{g}_t \in \mathcal{G}_{k,B}, N_{k,t}> N_{k,t-1}, N_{k,t-1}\in \Big(h_{k,n}(\Delta^{k,j-1}),h_{k,n}(\Delta^{k,j})\Big) \bigg\} \Delta^{k,j} \\
&\leq \sum_{j=1}^{B_k} \Big(h_{k,n}(\Delta^{k,j})-h_{k,n}(\Delta^{k,j-1})\Big) \Delta^{k,j} \\
\intertext{since $N_{k}$ can only be incremented a maximum of $h_{k,n}(\Delta^{k,j})-h_{k,n}(\Delta^{k,j-1})$ times while remaining in this range}
&= h_{k,n}(\Delta^{k,B_k})\Delta^{k,B_k} + \sum_{j=1}^{B_k-1}  h_{k,n}(\Delta^{k,j})\cdot(\Delta^{k,j}-\Delta^{k,j+1}) \\
&\leq h_{k,n}(\Delta^{k,B_k})\Delta^{k,B_k} + \int_{\Delta^{k,B_k}}^{\Delta^{k,1}} h_{k,n}(x)dx.
\end{align*} The last inequality holds since $h_{k,n}(x)$ are decreasing functions. $\square$

\section{Theorem \ref{thm::lower_bound} Proof: Lower bound on regret} \label{proof::lower_bound}
To prove Theorem \ref{thm::lower_bound}, we must define the additional quantities necessary to apply Theorem 1 of \cite{Graves1997} and frame the problem accordingly. 

We consider the reward history $(\mathbf{Y}_t)_{t=1}^n$ to be a realisation of a controlled Markov Chain moving on the state space $\mathbb{N}^K$ where the controls are the detection probability vectors selected in each round. Each control $\mathbf{g} \in \mathcal{G}$ then has an associated set of $\bs\lambda$ parameter vectors under which it is an optimal control $\Lambda_{\mathbf{g}} = \{\bs\lambda \in \mathbb{R}_{+}^K: \mathbf{g}^T\cdot \bs\lambda = \text{opt}_{\bs\lambda,\bs\gamma}\}$, which may be the empty set. For any states $\mathbf{y}, \mathbf{z} \in \mathbb{N}^K$ transition probabilities are straightforward Poisson probabilities due to independence across rounds:\begin{displaymath}
p(y,z;\bs\lambda,\mathbf{g}) = p(z;\bs\lambda,\mathbf{g})= \prod_{k=1}^K \frac{(g_k\lambda_k)^{z_k}e^{-g_k\lambda_k}}{z_k!}. 
\end{displaymath}  These transition probabilities define the Kullback Leibler Information number for any control $\mathbf{g} \in \mathcal{G}$: \begin{displaymath}
I^{\mathbf{g}}(\bs\lambda,\bs\theta)= \sum_{k=1}^K \log\bigg(\frac{p(z_k;\bs\lambda,\mathbf{g})}{p(z_k;\bs\theta,\mathbf{g})}\bigg)p(z_k;\bs\lambda,\mathbf{g}) =\sum_{k=1}^K \text{kl}(g_k\lambda_k,\gamma_k\theta_k) =\sum_{k=1}^K g_k \text{kl}(\lambda_k,\theta_k).
\end{displaymath}

With these quantities and those defined in Section \ref{sec::regret_lower_bound} we can apply Theorem 1 of \cite{Graves1997} to reach the following result for any uniformly good policy $\pi$
\begin{displaymath}
\lim \inf_{n \rightarrow \infty} \sum_{\mathbf{g} \in \mathcal{J}\setminus J(\bs\lambda)} \frac{I^{\mathbf{g}}(\bs\lambda,\bs\theta)\mathds{E}_{\bs\lambda}(\sum_{t=1}^n \mathds{I}\{\mathbf{g}_t = \mathbf{g}\})}{\log(n)} \geq 1 \text{ for every } \bs\theta \in B(\bs\lambda).
\end{displaymath}
Since $Reg_{\bs\lambda,\bs\gamma}^\pi(n)=\sum_{\mathbf{g} \in \mathcal{J}\setminus J(\bs\lambda)} \Delta_{\mathbf{g}}\mathds{E}_{\bs\lambda}(\sum_{t=1}^n \mathds{I}\{\mathbf{g}_t = \mathbf{g}\})$ the required result follows. $\square$
\clearpage

\section{Numerical Results} \label{app::tables}

\begin{table}[h!]
\centering \label{tab::test1}
\begin{tabular}{rrrrr}
  \hline
 Algorithm                            & Parameters & 0.025 Quantile & Median & 0.975 Quantile \\ 
  \hline
\multirow{6}{*}{FP-CUCB} & $\lambda_{max}=1$ & 9.52 & 11.96 & 15.89 \\ 
  & $\lambda_{max}=5$ & 36.42 & 42.53 & 50.03 \\ 
  & $\lambda_{max}=10$ & 57.44 & 72.57 & 88.70 \\ 
  & $\lambda_{max}=20$ & 89.07 & 117.97 & 143.95 \\ 
  & $\lambda_{max}=40$ & 123.23 & 178.07 & 223.81 \\ 
  & $\lambda_{max}=60$ & 143.87 & 215.46 & 276.25 \\ \hline
  \multirow{18}{*}{Thompson Sampling} & Mean=1, Variance=1 & 38.44 & 242.39 & 508.93 \\ 
  & Mean=5, Variance=1 & 1.95 & 132.79 & 358.15 \\ 
  & Mean=10, Variance=1 & 1.44 & 56.30 & 134.12 \\ 
  & Mean=20, Variance=1 & 11.66 & 17.76 & 25.88 \\ 
  & Mean=40, Variance=1 & 75.24 & 96.87 & 124.57 \\ 
  & Mean=60, Variance=1 & 122.72 & 180.67 & 233.25 \\ 
  & Mean=1, Variance=5 & 5.69 & 26.49 & 90.89 \\ 
  & Mean=5, Variance=5 & 2.32 & 38.51 & 134.07 \\ 
  & Mean=10, Variance=5 & 2.18 & 7.19 & 43.90 \\ 
  & Mean=20, Variance=5 & 7.17 & 10.95 & 15.80 \\ 
  & Mean=40, Variance=5 & 30.00 & 36.11 & 43.23 \\ 
  & Mean=60, Variance=5 & 57.61 & 72.42 & 87.30 \\ 
  & Mean=1, Variance=10 & 6.31 & 14.21 & 36.57 \\ 
  & Mean=5, Variance=10 & 3.60 & 9.35 & 35.87 \\ 
  & Mean=10, Variance=10 & 3.28 & 6.65 & 18.41 \\ 
  & Mean=20, Variance=10 & 6.55 & 9.67 & 15.97 \\ 
  & Mean=40, Variance=10 & 20.15 & 24.65 & 30.25 \\ 
  & Mean=60, Variance=10 & 40.17 & 46.12 & 55.09 \\ \hline
  Greedy & & 79.77 & 679.76 & 1657.52 \\ 
   \hline
\end{tabular}
\caption{Quantiles of scaled regret at horizon $n=2000$ for algorithms applied to Test (i) data}
\end{table}
\clearpage

\begin{table}[h!]
\centering \label{tab::test2}
\begin{tabular}{rrrrr}
 \hline
 Algorithm                            & Parameters & 0.025 Quantile & Median & 0.975 Quantile \\ 
  \hline
\multirow{6}{*}{FP-CUCB} &$\lambda_{max}=1$ & 65.82 & 75.92 & 89.31 \\ 
  &$\lambda_{max}=5$ & 269.77 & 297.74 & 323.55 \\ 
  &$\lambda_{max}=10$ & 433.92 & 480.78 & 517.22 \\ 
  &$\lambda_{max}=20$ & 577.74 & 661.08 & 754.25 \\ 
  &$\lambda_{max}=40$ & 643.61 & 759.90 & 891.13 \\ 
  &$\lambda_{max}=60$ & 665.45 & 794.38 & 931.36 \\ \hline
  \multirow{18}{*}{Thompson Sampling} & Mean=1, Variance=1 & 286.11 & 603.51 & 969.56 \\ 
  & Mean=5, Variance=1 & 7.94 & 184.48 & 568.05 \\ 
  & Mean=10, Variance=1 & 8.12 & 21.05 & 159.10 \\ 
  & Mean=20, Variance=1 & 102.40 & 132.00 & 174.17 \\ 
  & Mean=40, Variance=1 & 286.61 & 395.04 & 472.38 \\ 
  & Mean=60, Variance=1 & 371.53 & 504.06 & 609.47 \\ 
  & Mean=1, Variance=5 & 26.95 & 61.18 & 153.86 \\ 
  & Mean=5, Variance=5 & 9.56 & 70.19 & 224.13 \\ 
  & Mean=10, Variance=5 & 6.55 & 13.80 & 40.48 \\ 
  & Mean=20, Variance=5 & 36.57 & 45.19 & 56.15 \\ 
  & Mean=40, Variance=5 & 128.60 & 172.27 & 208.23 \\ 
  & Mean=60, Variance=5 & 222.33 & 303.67 & 361.93 \\ 
  & Mean=1, Variance=10 & 25.38 & 41.44 & 69.92 \\ 
  & Mean=5, Variance=10 & 12.61 & 26.23 & 100.81 \\ 
  & Mean=10, Variance=10 & 10.22 & 15.79 & 32.32 \\ 
  & Mean=20, Variance=10 & 24.28 & 30.60 & 39.17 \\ 
  & Mean=40, Variance=10 & 84.45 & 106.17 & 122.09 \\ 
  & Mean=60, Variance=10 & 151.68 & 206.13 & 244.60 \\ \hline
  Greedy & & 296.46 & 720.45 & 1163.15 \\ 
   \hline
\end{tabular}
\caption{Quantiles of scaled regret at horizon $n=2000$ for algorithms applied to Test (ii) data}
\end{table}

\clearpage

\begin{table}[h!]
\centering \label{tab::test3}
\begin{tabular}{rrrrr}
  \hline
 Algorithm & Parameter & 0.025 Quantile & Median & 0.975 Quantile \\ 
  \hline
\multirow{6}{*}{FP-CUCB} & $\lambda_{max}=1$ & 2.17 & 3.37 & 7.20 \\ 
  & $\lambda_{max}=10$ & 9.19 & 10.34 & 11.78 \\ 
  & $\lambda_{max}=25$ & 15.92 & 18.45 & 21.33 \\ 
  & $\lambda_{max}=50$ & 22.57 & 27.39 & 31.58 \\ 
  & $\lambda_{max}=100$ & 30.41 & 37.59 & 44.85 \\ 
  & $\lambda_{max}=200$ & 38.90 & 48.07 & 57.96 \\ \hline
  \multirow{18}{*}{Thompson Sampling} & Mean=1, Variance=5 & 30.47 & 66.95 & 115.65 \\ 
  & Mean=10, Variance=5 & 36.78 & 64.41 & 98.24 \\ 
  & Mean=25, Variance=5 & 29.06 & 58.44 & 95.57 \\ 
  & Mean=50, Variance=5 & 10.82 & 39.65 & 71.71 \\ 
  & Mean=100, Variance=5 & 4.83 & 6.05 & 7.71 \\ 
  & Mean=200, Variance=5 & 28.24 & 34.20 & 40.37 \\ 
  & Mean=1, Variance=10 & 12.61 & 52.06 & 97.08 \\ 
  & Mean=10, Variance=10 & 33.99 & 68.30 & 109.44 \\ 
  & Mean=25, Variance=10 & 30.97 & 64.55 & 105.03 \\ 
  & Mean=50, Variance=10 & 17.32 & 46.39 & 80.35 \\ 
  & Mean=100, Variance=10 & 4.26 & 5.52 & 7.09 \\ 
  & Mean=200, Variance=10 & 21.37 & 25.06 & 29.00 \\ 
  & Mean=1, Variance=25 & 3.87 & 37.19 & 102.98 \\ 
  & Mean=10, Variance=25 & 36.51 & 66.12 & 107.72 \\ 
  & Mean=25, Variance=25 & 30.87 & 64.73 & 106.71 \\ 
  & Mean=50, Variance=25 & 20.21 & 51.32 & 86.70 \\ 
  & Mean=100, Variance=25 & 3.86 & 5.09 & 6.79 \\ 
  & Mean=200, Variance=25 & 14.08 & 15.92 & 18.09 \\ \hline
  Greedy & & 21.57 & 49.20 & 95.89 \\ 
   \hline
\end{tabular} 
\caption{Quantiles of scaled regret at horizon $n=2000$ for algorithms applied to Test (iii) data}
\end{table}

\clearpage

\begin{table}[h!]
\centering \label{tab::test4}
\begin{tabular}{rrrrr}
  \hline
 Algorithm & Parameters & 0.025 Quantile & Median & 0.975 Quantile \\ 
  \hline
\multirow{6}{*}{FP-CUCB} & $\lambda_{max}=0.1$ & 47.56 & 87.07 & 162.36 \\ 
  & $\lambda_{max}=1$ & 62.60 & 108.48 & 195.80 \\ 
  & $\lambda_{max}=5$ & 98.70 & 163.13 & 279.62 \\ 
  & $\lambda_{max}=10$ & 109.59 & 184.01 & 311.37 \\ 
  & $\lambda_{max}=20$ & 116.40 & 200.99 & 336.25 \\ 
  & $\lambda_{max}=40$ & 120.39 & 210.65 & 356.25 \\ \hline 
  \multirow{18}{*}{Thompson Sampling} & Mean=0.1, Variance=1 & 70.68 & 136.84 & 284.14 \\ 
  & Mean=1, Variance=1 & 42.78 & 61.44 & 91.98 \\ 
  & Mean=5, Variance=1 & 43.96 & 75.38 & 119.21 \\ 
  & Mean=10, Variance=1 & 75.45 & 118.86 & 197.36 \\ 
  & Mean=20, Variance=1 & 104.58 & 174.02 & 291.32 \\ 
  & Mean=40, Variance=1 & 119.72 & 207.46 & 349.74 \\ 
  & Mean=0.1, Variance=5 & 94.23 & 246.71 & 467.06 \\ 
  & Mean=1, Variance=5 & 43.48 & 73.41 & 119.94 \\ 
  & Mean=5, Variance=5 & 41.71 & 60.07 & 88.64 \\ 
  & Mean=10, Variance=5 & 45.15 & 72.69 & 119.42 \\ 
  & Mean=20, Variance=5 & 69.43 & 113.12 & 191.90 \\ 
  & Mean=40, Variance=5 & 102.60 & 169.98 & 281.94 \\ 
  & Mean=0.1, Variance=10 & 134.60 & 320.63 & 588.63 \\ 
  & Mean=1, Variance=10 & 48.26 & 81.35 & 146.95 \\ 
  & Mean=5, Variance=10 & 41.43 & 58.66 & 84.74 \\ 
  & Mean=10, Variance=10 & 40.78 & 62.10 & 99.55 \\ 
  & Mean=20, Variance=10 & 55.42 & 89.68 & 146.88 \\ 
  & Mean=40, Variance=10 & 86.98 & 141.99 & 239.18 \\ \hline
  Greedy &  & 664.28 & 1825.61 & 1999.89 \\ 
   \hline
\end{tabular}

\caption{Quantiles of scaled regret at horizon $n=2000$ for algorithms applied to Test (iv) data}
\end{table}

\end{document}